\documentclass[letterpaper]{article} 
\usepackage{aaai2026}
\usepackage{times}  
\usepackage{helvet}  
\usepackage{courier}  
\usepackage[hyphens]{url}  
\usepackage{graphicx} 
\usepackage{natbib}  
\usepackage{caption} 
\usepackage{booktabs}
\usepackage{algorithm}
\usepackage{algorithmic}

\usepackage{newfloat}
\usepackage{listings}
\DeclareCaptionStyle{ruled}{labelfont=normalfont,labelsep=colon,strut=off} 
\floatstyle{ruled}
\newfloat{listing}{tb}{lst}{}
\floatname{listing}{Listing}

\title{Assessing the Capabilities of LLMs in Humor: \\ A Multi-dimensional Analysis of Oogiri Generation and Evaluation}

\author{
    Ritsu Sakabe ${}^{1}$ Hwichan Kim ${}^{1,2}$ Tosho Hirasawa ${}^{1,2}$ Mamoru Komachi ${}^{1}$
}
\affiliations{
    \textsuperscript{\rm 1}Hitotsubashi University\\ \textsuperscript{\rm 2}Tokyo Metropolitan University

    dm240011@g.hit-u.ac.jp, kim@edu.sds.hit-u.ac.jp, toshosan@tmu.ac.jp, mamoru.komachi@r.hit-u.ac.jp
}

\usepackage{bibentry}

\begin{document}

\maketitle

\begin{abstract}
Computational humor is a frontier for creating advanced and engaging natural language processing (NLP) applications, such as sophisticated dialogue systems. 
While previous studies have benchmarked the humor capabilities of Large Language Models (LLMs), they have often relied on single-dimensional evaluations, such as judging whether something is simply ``funny.'' 
This paper argues that a multifaceted understanding of humor is necessary and addresses this gap by systematically evaluating LLMs through the lens of Oogiri, a form of Japanese improvisational comedy games.
To achieve this, we expanded upon existing Oogiri datasets with data from new sources and then augmented the collection with Oogiri responses generated by LLMs. 
We then manually annotated this expanded collection with 5-point absolute ratings across six dimensions: Novelty, Clarity, Relevance, Intelligence, Empathy, and Overall Funniness. 
Using this dataset, we assessed the capabilities of state-of-the-art LLMs on two core tasks: their ability to generate creative Oogiri responses and their ability to evaluate the funniness of responses using a six-dimensional evaluation. 
Our results show that while LLMs can generate responses at a level between low- and mid-tier human performance, they exhibit a notable lack of Empathy. 
This deficit in Empathy helps explain their failure to replicate human humor assessment. 
Correlation analyses of human and model evaluation data further reveal a fundamental divergence in evaluation criteria: LLMs prioritize Novelty, whereas humans prioritize Empathy. 
We release our annotated corpus to the community to pave the way for the development of more emotionally intelligent and sophisticated conversational agents.
\end{abstract}


\section{Introduction}
\begin{figure}[t]
    \setlength\abovecaptionskip{2truemm}
    \centering
    \includegraphics[width=1\linewidth]{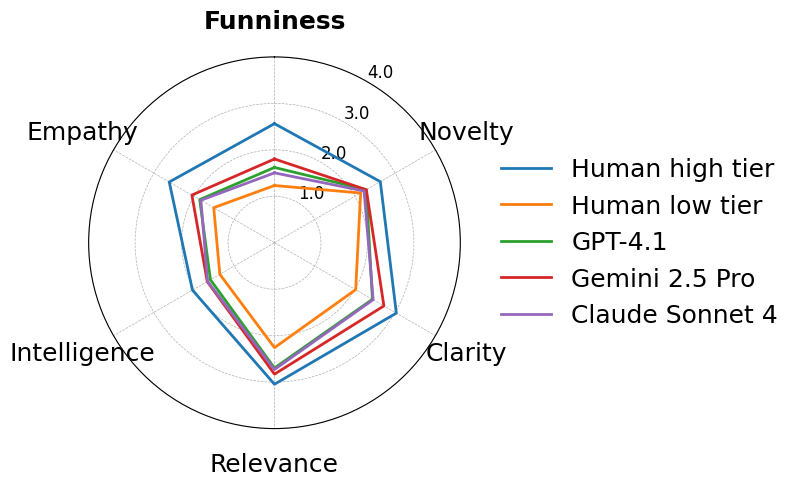}
    \caption{Multi-dimensional comparison of Oogiri responses from humans and LLMs, as evaluated by humans. ``Human high tier'' and ``human low tier'' refer to the highest- and lowest-rated human responses, as determined by manual annotation. (Funniness refers to Overall Funniness.)}
    \label{fig:Radar_Chart}
\end{figure}

Humor is a crucial aspect of human cognition: it facilitates social bonding \cite{doi:10.1152/advan.00030.2017} and correlates with general intelligence \cite{greengross2011humor}.
In an effort to realize this complex and human activity with AI, recent studies have begun to assess the humor comprehension, evaluation, and generation abilities of large language models (LLMs)~\cite{hessel-etal-2023-androids, romanowski-etal-2025-punchlines, chen-etal-2024-u}.
However, much of this existing research focuses on the ``performance improvement'' aspect---that is, how to make LLMs generate or evaluate more humorous responses. 
On the other hand, there has been insufficient analysis of how the structure of funniness as perceived by LLMs fundamentally differs from that of humans---in other words, the qualitative differences in their ``humor sensibilities.''
Understanding this difference in sensibilities is essential for discerning whether LLMs interpret humor in the same way as humans, or if they possess their own unique preferences for what constitutes humor. 

\begin{figure*}[t]
    \setlength\abovecaptionskip{2truemm}
    \centering
    \includegraphics[width=1.0\linewidth]{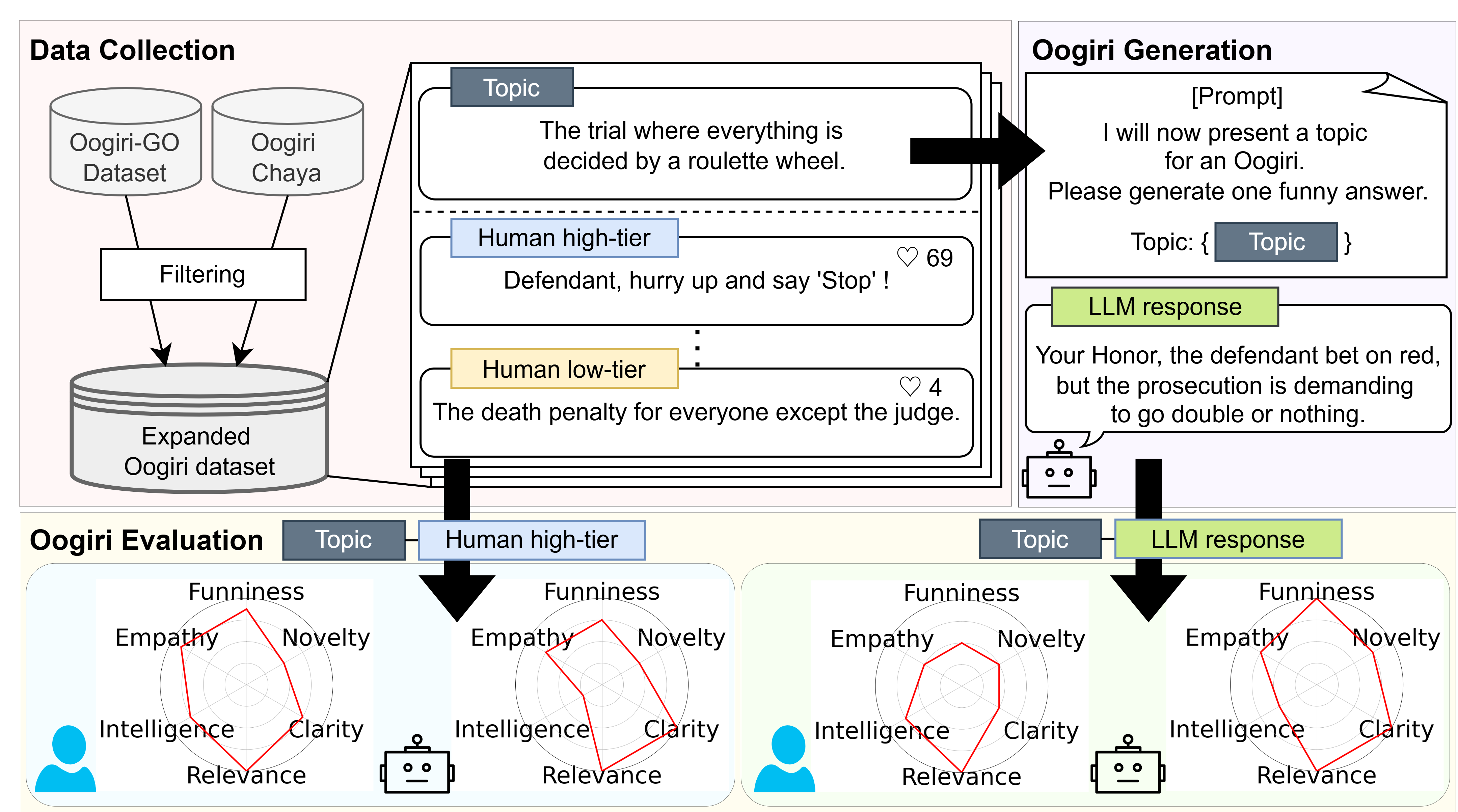}
    \caption{Overview of the experimental process for Oogiri generation and evaluation.}
    \label{fig:experiment_flow}
\end{figure*}

To bridge this sensibility gap and create more natural and engaging conversational agents, two fundamental abilities must be addressed. 
The first is humor generation (RQ1): the ability to produce contextually appropriate and creative humor. 
Developing this ability, however, relies on robust evaluation methods. 
While human assessment is the most trusted benchmark, the recent ``LLM-as-a-Judge'' paradigm offers a scalable alternative. 
This gives rise to the second, equally critical challenge: humor evaluation (RQ2). 
Before we can trust LLMs to guide the development of humorous AI, we must first understand if they can evaluate humor with human-like criteria and sensibilities. 

In this study, we conduct detailed analyses of the humor capabilities of state-of-the-art LLMs such as GPT-4.1.
We begin by collecting datasets of ``Oogiri,'' a uniquely Japanese comedic format in which participants are given a topic (e.g., `A taxi driver who has awakened') and respond with a humorous punchline (e.g., `I can make this traffic jam ``disappear.'' Should I?').
Our primary resource is the Oogiri-GO dataset \cite{zhong2024letsthinkoutsidebox}, collected from Bokete, an online Oogiri competition platform.
To broaden the range of topics and human-written responses, we additionally collect data from another platform, ``Oogiri-Chaya.''
Using these resources, we address two core research questions regarding the humor capabilities of LLMs (Figure \ref{fig:experiment_flow}).

Our first research question asks: \textit{Can LLMs produce humorous responses that humans judge as funny?}
To address RQ1, we prompted several LLMs with the collected topics and then evaluated both the LLM-generated and human-generated responses through crowd annotations.
Previous studies have mainly relied on binary judgments of whether a response is funny \cite{romanowski-etal-2025-punchlines, zhong2024letsthinkoutsidebox}, but we argue that funniness is multifaceted, encompassing relevance, empathy, and other qualities \cite{hampes2001relation}.
We therefore conducted a manual annotation along six dimensions: Novelty, Clarity, Relevance, Empathy, Intelligence, and Overall Funniness.
Each dimension is rated on a 0--4 Likert scale (0 = not at all, 4 = excellent).
This fine-grained evaluation allows us to determine not only whether LLMs can produce humor, but also which specific dimensions they excel at or struggle with.

Our second research question asks: \textit{Can LLMs judge humorous responses that humans judge as funny?}
To address RQ2, we provide the LLMs with each topic--response pair and instruct them to perform an evaluation using the same six-dimensional rubric described previously. 
We then compute the correlation between the human annotations and LLMs' scores.
These correlation coefficients allow us to verify how reliably LLMs can assess Oogiri responses compared to human judges, and which evaluation dimensions LLMs prioritize when evaluating humor.

In summary, our contributions are three-fold:

\paragraph{Creation of a Multi-Dimensional Annotated Corpus for Oogiri Humor Research:} 
To provide a reliable resource for humor research, we constructed a new, large-scale annotated corpus. 
The core of this contribution lies in the novel annotation of this data collection by both human annotators and the latest LLMs, applying 5-point ratings across our six proposed dimensions (Figure \ref{fig:Radar_Chart}). 
We are releasing these annotations and our benchmark design to the community to provide a reliable resource for multifaceted humor analysis.

\paragraph{Assessment of State-of-the-Art LLMs' Humor Generation Capabilities:} 
We multi-dimensionally analyze the humor generation abilities of state-of-the-art LLMs (GPT-4.1, Gemini 2.5 Pro, Claude Sonnet 4) using manual, six-dimensional scoring of their generated responses.
Our findings show that LLMs possess an Oogiri generation ability equivalent to that of low- to mid-tier humans, with Gemini 2.5 Pro being the most proficient among the models tested (Figure \ref{fig:Radar_Chart}).
An analysis across the evaluation dimensions reveals that the primary weakness of LLMs compared to humans is on the Empathy dimension. 

\paragraph{Assessment of State-of-the-Art LLMs' Humor Evaluation Capabilities:} 
By comparing response evaluations across six dimensions from both humans and LLMs, we assess the degree of alignment between the two. 
This allows us to determine the extent to which LLMs can serve as proxies for human humor judges and to clarify which factors LLMs prioritize when evaluating humor.
Our analysis reveals that while Claude Sonnet 4 showed the highest correlation with human evaluations among the models, overall agreement between humans and LLMs is limited. 
Furthermore, mirroring a tendency observed in the generation results, we found that while humans prioritize the Empathy dimension most when evaluating humor, LLMs prioritize the Novelty dimension. 
This finding suggests that LLMs may possess their unique ``sense of humor,'' highlighting the complexity of aligning an AI's humor sensibilities with those of humans.

\section{Related Work}
\subsection{Computational Humor}
Humor has long been studied as a challenging aspect of AI and NLP. 
Early theories suggest that humor creation and appreciation require advanced cognitive abilities, such as the ability to recognize incongruities and shared context.
For instance, a prominent cognitive framework, \citet{Suls1972} two-stage model, posits that humor is appreciated when a perceiver first detects an incongruity in a joke and then successfully resolves it by finding a cognitive rule that reconciles the punchline with the setup.
Empirical research by ~\citet{greengross2011humor} even showed that humor ability is correlated with human intelligence and mating success, implying that humor might be an inherent component of what we consider ``intelligence.'' 
In the context of AI, humor understanding and generation test a model's semantic and pragmatic knowledge, creativity, and cultural awareness. 

Recent NLP studies, using materials like jokes, puns, and satirical headlines, have tackled tasks such as humor detection (determining if a given text is humorous) and humor understanding/explanation (identifying why something is funny or choosing the correct punchline for a setup).
~\citet{baranov-etal-2023-told} examined the robustness of humor detection models and found that their performance can degrade when jokes are phrased differently or in new domains, suggesting that humor detection systems lack true general understanding. 
~\citet{hessel-etal-2023-androids} introduced a benchmark from the New Yorker Caption Contest to evaluate humor understanding by having models match cartoons with the funniest caption; even strong language models struggled with this task, often failing to predict which caption humans found funniest. 
In short, despite the impressive fluency of modern LLMs, truly ``getting'' humor remains an open challenge in AI.

\subsection{Humor Generation}
In the realm of humor generation, the challenge for AI is even more pronounced. 
Producing original humor has proven extremely difficult, as jokes often require creativity, world knowledge, and a fine sense of timing or surprise. For instance, ~\citet{jentzsch-kersting-2023-chatgpt} demonstrated that while models like ChatGPT can be amusing in conversation, they frequently miss the mark on humor.
Their outputs tend to be formulaic or overly safe, lacking the spark that makes humans laugh. 
Several other studies have specifically addressed this difficulty.

One interesting approach by ~\citet{horvitz2024gettinghumorcraftinghumor} involved using LLMs to manipulate humor in text.
They showed that an LLM can effectively remove the humor from a satirical news headline (making it ``unfunny''), but doing the reverse (adding humor to a serious headline or creating a new joke) is much more challenging. 
This asymmetry indicates that LLMs, while good at understanding literal meaning, struggle with the creative leap required for humor. 
The limitations of LLM humor generation have been observed across multiple studies: models often rely on puns, clichés, or shallow wordplay and may fail to surprise the audience---an essential element of humor. 
These findings underscore that humor generation is a high-bar task pushing the boundaries of what LLMs can do, and improvements likely require advances in knowledge representation and perhaps incorporating cultural or common-sense reasoning.

\subsection{Humor Evaluation}
With the increasing capabilities of LLMs, a paradigm known as ``LLM-as-a-Judge,'' which uses LLMs themselves as evaluators to substitute for human assessment, has garnered significant attention. 
Various attempts have also been made in humor research.

\citet{romanowski-etal-2025-punchlines} have proposed a new metric that involves transcribing stand-up comedy, inputting it into an LLM, and evaluating whether the model can accurately extract the funny parts. 
Their evaluation of major LLMs shows that the accuracy remains around 50 \%. It is also reported that various prompting techniques, such as assigning a persona, do not lead to notable improvements.

Furthermore, \citet{hwang-etal-2025-bottlehumor} have proposed an LLM-based method to understand and explain the funniness of multimodal humor (e.g., memes and cartoons) composed of images and captions. 
Their method uses the information bottleneck principle to extract relevant knowledge from a Vision-Language Model (VLM) and iteratively refine it. 
This research highlights the importance of world knowledge in humor comprehension.

\subsection{Oogiri}
The Oogiri format of humor has gained attention as a challenging test of creativity. 
\citet{Nakagawa2019Analysis} conducted one of the first analyses of Oogiri from a computational perspective. 
They used crowdsourcing to annotate Oogiri responses with various humor-related attributes and found that three factors (relevance to the topic, clarity, and novelty) were the most important contributors to a response's funniness. 
This suggests that a good Oogiri response must be on-topic (relevant), easy to understand, yet offer a fresh or unexpected twist. 

More recently, the Oogiri-GO dataset by ~\citet{zhong2024letsthinkoutsidebox} provided a large-scale resource for multilingual humor research.
The dataset compiled Oogiri topics and responses from the web and was used to benchmark the humor generation and evaluation abilities of LLMs. 
As mentioned earlier, their results showed that the models performed poorly in both generating humorous responses and comparatively evaluating humor. 
Our work builds on these insights by focusing specifically on Japanese-language humor and addressing some limitations of Oogiri-GO. 

\section{Dataset}
This section describes the sources and curation process for our Oogiri dataset. 
The resulting collection of topics and responses is the foundation for the multi-dimensional manual annotation process, detailed in the Methodology section.

\subsection{Sources of Oogiri Samples}
\label{subsec:data_sources}
To assemble a diverse collection of human-written Oogiri topics and responses, we drew on the two distinct sources.

\paragraph{Oogiri-GO} is a dataset proposed by \citet{zhong2024letsthinkoutsidebox}.
This dataset is derived from Bokete\footnote{\url{https://bokete.jp/}}, an online Oogiri competition platform.  
Each sample comprises three elements: a topic, a response submitted by a human competitor, and the number of votes that the response received from human judges based on its funniness.
The topics are grouped into two categories: text-based and image-based.  
Text-based topics are expressed entirely in natural language, whereas image-based topics present an image for which participants must craft a humorous caption.
\footnote{
Image-based topics consisting solely of text were transcribed using GPT-4.1 and reclassified as text-based.
}

\paragraph{Oogiri-Chaya} is a dataset derived from another online Oogiri competition platform\footnote{\url{https://oogiri-chaya.com/}}.
We collected all samples between 21 November 2021 and 17 July 2024.  
The dataset mirrors the structure of Oogiri-GO: each entry contains a topic, a participant's response, and the number of votes that response received, and is labelled as either text-based or image-based.  
After collecting samples, we detected topics and responses that potentially contain discriminatory, violent, sexual, or otherwise socially inappropriate content using GPT-4o. Then, we manually reviewed all flagged items and removed those deemed inappropriate.

\paragraph{Characteristics of the two Oogiri sources}
The two data sources differ significantly in terms of their evaluation systems and the potential biases that may arise from these systems. 
In Bokete, the data source for Oogiri-GO, the design allows users to post and vote while viewing other users' responses and their current vote counts. 
This system can potentially give rise to a ``first-mover advantage,'' where earlier responses are favored.
It can also cause a ``conformity bias'' (bandwagon effect), in which popular responses tend to attract more votes. 
Consequently, there is a concern that the vote counts may be influenced not only by the pure funniness of a response but also by the timing of submission and other users' evaluations.

On the other hand, Oogiri-Chaya is designed to systematically eliminate these biases. 
During the submission period, no other participants' responses are visible, and during the evaluation period, the vote counts for each response remain hidden. 
Therefore, participants can evaluate responses based on pure funniness without being influenced by others, making the evaluation data from Oogiri-Chaya a more objective and reliable indicator of funniness compared to Bokete. 
In particular, responses with low vote counts can be considered to have been judged as ``unfunny'' with a high degree of certainty, free from the influence of such biases.
By incorporating samples from Oogiri-Chaya, we not only broaden the range of topics and responses but also enhance the overall reliability of our dataset with less biased evaluation data.

\subsection{Ensuring the Quality of Samples}
To guarantee the quality of the dataset, we applied filtering procedures to both the Oogiri-GO and Oogiri-Chaya data sources, selecting only samples that met specific participation and voting criteria.

For the Oogiri-GO dataset, we retained only those topics that satisfied two conditions: (1) the top-voted response had at least 100 votes, and (2) the topic had at least 10 submitted responses. 
This filtering process resulted in a final set of 1,329 topics (313 text-based and 1,016 image-based).

Similarly, for the Oogiri-Chaya dataset, we selected only topics that had 100 or more responses, yielding a total of 551 topics (425 text-based and 126 image-based).

\begin{table*}[t]
    \centering
    \begin{tabular}{l|rrrrrr}
        \toprule
        & {\textbf{Novelty}} & {\textbf{Clarity}} & {\textbf{Relevance}} & {\textbf{Intelligence}} & {\textbf{Empathy}} & {\textbf{Overall Funniness}} \\
        \midrule
        High & \textbf{2.630} (0.578) & \textbf{3.029} (0.691) & \textbf{3.046} (0.665) & \textbf{2.039} (0.708) & \textbf{2.615} (0.656) & \textbf{2.563} (0.648) \\
        Mid & 2.320 (0.621) & 2.571 (0.822) & 2.652 (0.724) & 1.732 (0.669) & 2.100 (0.715) & 1.913 (0.675) \\
        Low & 2.142 (0.690) & 2.018 (0.928) & 2.258 (0.876) & 1.358 (0.614) & 1.510 (0.738) & 1.232 (0.658) \\
        Unrelated & 2.377 (0.630) & 0.898 (0.702) & 0.935 (0.727) & 1.032 (0.613) & 0.725 (0.615) & 0.681 (0.591) \\
        \midrule
        GPT-4.1 (serious) & 1.369 (0.826) & 2.836 (0.728) & \underline{3.055} (0.623) & 1.786 (0.741) & 2.023 (0.710) & 0.976 (0.654) \\
        GPT-4.1 & 2.282 (0.608) & 2.435 (0.840) & 2.696 (0.691) & 1.592 (0.656) & 1.858 (0.770) & 1.621 (0.689) \\
        Gemini 2.5 Pro & 2.283 (0.616) & 2.719 (0.734) & 2.828 (0.674) & 1.671 (0.638) & 2.053 (0.700) & 1.803 (0.700) \\
        Claude Sonnet 4 & 2.227 (0.597) & 2.452 (0.815) & 2.729 (0.711) & 1.662 (0.636) & 1.830 (0.707) & 1.504 (0.669) \\
        \bottomrule
    \end{tabular}
    \caption{Mean (and standard deviation) of human evaluations of Oogiri generation.}
    \label{tab:Human_reordered}
\end{table*}

\section{Methodology}
\label{Methodology}
This section details the overall experimental design used to address our research questions.
For RQ1, we investigate whether state-of-the-art LLMs can produce Oogiri responses that humans perceive as funny, and how their humor compares with that of humans.  
Because Oogiri humor is inherently multidimensional, assessing it along a single funniness dimension is insufficient.  
To capture its nuances, we evaluate each response across several complementary dimensions.  
We let the LLMs generate responses for each Oogiri topic and then ask human annotators to rate each response along these dimensions.  
The resulting scores reveal the strengths and limitations of current LLMs in humorous response generation.
For RQ2, we investigate whether state-of-the-art LLMs can evaluate Oogiri responses that humans perceive as funny, and to what extent the characteristics of their humor evaluation align with those of humans. 

\subsection{Experimental Materials and Stimuli}
To analyze the characteristics of Oogiri generation, we instructed the latest models---GPT-4.1 (gpt-4.1-2025-04-14), Gemini 2.5 Pro (gemini-2.5-pro-preview-05-06), and Claude Sonnet 4 (claude-sonnet-4-20250514-v1:0)---to generate Oogiri responses. 
For each Oogiri topic, we embedded it within a base prompt (see Appendix A for details) written in Japanese to solicit a humorous response.
All generation was performed with default parameter settings. 

From the collected datasets (Oogiri-GO and Oogiri-Chaya), we selected 200 topics (100 text-based and 100 image-based) based on the popularity of their top-voted answers. 
For each topic, we prepared a final evaluation set consisting of the following eight types of responses:

\begin{itemize}
    \item \textbf{High/Mid/Low}: The response with the most/average/fewest votes on the source website
    \item \textbf{Unrelated}: The top-voted response from a different, irrelevant topic
    \item \textbf{GPT-4.1 (serious)}: A serious (non-humorous) response by GPT-4.1
    \item \textbf{GPT-4.1}: A (humorous) response by GPT-4.1
    \item \textbf{Gemini 2.5 Pro}: A response by Gemini 2.5 Pro
    \item \textbf{Claude Sonnet 4}: A response by Claude Sonnet 4
\end{itemize}

\subsection{Evaluation Procedure}
\paragraph{Multi-dimensional Evaluation Framework} 
We conducted absolute evaluations for the eight response types across six dimensions: Novelty, Clarity, Relevance, Intelligence, Empathy, and Overall Funniness. 
Each dimension was rated on a 0--4 integer scale. 
The specific definitions provided to annotators are as follows:

\begin{enumerate}
    \item \textbf{Novelty}: Measures the originality of the response to the topic. A score of 0 means ``not new at all (something anyone could think of),'' while a 4 means ``very new (a unique perspective that others would not think of).''
    \item \textbf{Clarity}: Measures how easy the response is to understand. A 0 indicates ``completely incomprehensible (the meaning is unclear even after multiple readings),'' and a 4 indicates ``easily comprehensible (the meaning is immediately clear).''
    \item \textbf{Relevance}: Measures the connection between the topic and the response. A 0 means ``completely unrelated to the topic,'' while a 4 means ``very closely related to the topic.''
    \item \textbf{Intelligence}: Measures the intellectual quality of the response. A 0 means ``contains no intellectual elements (silly or trivial),'' and a 4 means ``highly intelligent (conveys intellect or sophistication).''
    \item \textbf{Empathy}: Measures how relatable the response is. A 0 means ``not relatable at all (unconvincing, hard to imagine the situation or feelings),'' while a 4 means ``highly relatable (the situation and feelings are well understood).''
    \item \textbf{Overall Funniness}: Measures the pure humor of the response, integrating all dimensions and factors. A 0 means ``not funny at all,'' and a 4 means ``extremely funny.''
\end{enumerate}

\paragraph{Human and LLM Annotation Process}
Both human and LLM annotators performed the evaluation. 
For human annotation, we used Lancers\footnote{https://www.lancers.jp}. 
Each response was rated by four different native Japanese speakers, recruited without regard to age or educational background. 
To ensure a fair rating, we created eight distinct subsets of the data, ensuring each annotator evaluated only one response type per topic. 
For LLM annotation, we used the same models from the generation task (GPT-4.1, Gemini 2.5 Pro, Claude Sonnet 4) and provided them with the same instructions and evaluation framework as the human annotators.
The prompt template for the LLM evaluation task is shown in Appendix A.

\begin{figure}[t]
    \setlength\abovecaptionskip{2truemm}
    \centering
    \includegraphics[width=0.9\linewidth]{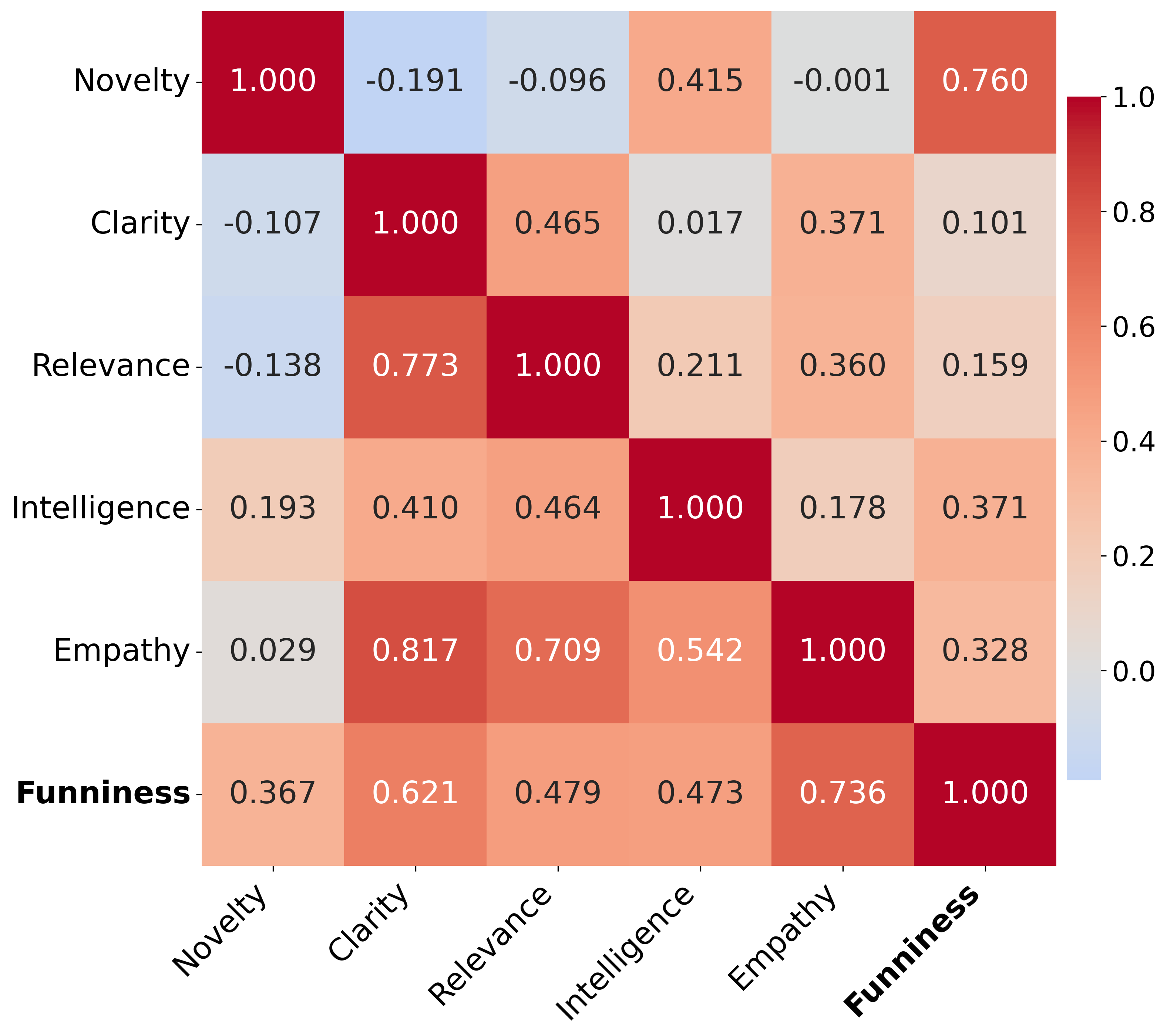}
    \caption{Spearman correlation matrix of human and GPT-4.1 evaluation. Human evaluations are in the lower triangle, and GPT-4.1 evaluations are in the upper triangle. GPT-4.1 is a representative example; see Appendix E for other models.}
    \label{fig:dimension_Heat_Human_and_GPT}
\end{figure}

\begin{table}[t]
\centering
\begin{tabular}{p{5.6cm}rr}
\toprule
\textbf{Responses} & \textbf{Emp.} & \textbf{Fun.} \\
\midrule
A funeral director shouldn't say things like, `See you next time!' (High) & 3.00 & 2.00 \\
\midrule
So, this makes it zero calories now, right? (GPT-4.1) & 1.25 & 0.75 \\
\bottomrule
\end{tabular}
\caption{Examples for the topic ``Tear a coupon to pieces and say one line.'' illustrating discrepancies in human ratings of empathy and funniness between human and LLM responses.}
\label{tab:example_RQ1}
\end{table}


\subsection{Analysis Procedure}
For RQ1, we analyze the human evaluation scores for each response type. 
For RQ2, we compute the Spearman's correlation coefficient between the human scores and the LLM scores. 
A high correlation would indicate that an LLM can mimic human evaluation abilities, while a low correlation would suggest a divergence in their evaluation criteria.

\begin{table*}[t]
    \centering
    
    \begin{tabular}{l|cccccc}
        \toprule
        Model            & \textbf{Novelty} & \textbf{Clarity} & \textbf{Relevance} & \textbf{Intelligence} & \textbf{Empathy} & \textbf{Overall Funniness} \\
        \midrule
        GPT-4.1          & 0.361          & 0.379          & \textbf{0.441} & \textbf{0.352} & \textbf{0.303} & 0.224 \\
        Gemini 2.5 Pro   & \textbf{0.393} & 0.356          & 0.328          & 0.188          & 0.248          & 0.169 \\
        Claude Sonnet 4  & 0.375          & \textbf{0.383} & 0.315          & 0.275          & 0.278          & \textbf{0.266} \\
        \bottomrule
    \end{tabular}
    \caption{Spearman correlation coefficients of Oogiri evaluation between humans and LLMs.}
    \label{tab:spearman_summary}
\end{table*}

\begin{table*}[t]
    \centering
    \begin{tabular}{l|rrrrrr}
        \toprule
        & {\textbf{Novelty}} & {\textbf{Clarity}} & {\textbf{Relevance}} & {\textbf{Intelligence}} & {\textbf{Empathy}} & {\textbf{Overall Funniness}} \\
        \midrule
        High & 2.416 (0.884) & 3.877 (0.445) & 3.719 (0.674) & 1.466 (0.884) & 2.602 (0.841) & 2.637 (0.787) \\
        Mid & 2.376 (0.924) & 3.875 (0.448) & 3.652 (0.717) & 1.416 (0.852) & 2.617 (0.842) & 2.604 (0.785) \\
        Low & 2.429 (0.953) & 3.702 (0.645) & 3.514 (0.832) & 1.376 (0.888) & 2.286 (0.856) & 2.461 (0.813) \\
        Unrelated & \textbf{2.722} (0.964) & 3.158 (1.072) & 2.624 (1.065) & 1.333 (0.872) & 2.075 (0.987) & 2.411 (0.863) \\
        \midrule
        GPT-4.1 (serious) & 1.195 (1.113) & \underline{3.997} (0.050) & \underline{3.932} (0.359) & 1.659 (0.940) & 2.749 (0.861) & 1.404 (1.125) \\
        GPT-4.1 & 2.524 (0.712) & 3.987 (0.132) & \textbf{3.917} (0.348) & 1.627 (0.675) & \textbf{2.955} (0.739) & 2.895 (0.629) \\
        Gemini 2.5 Pro & 2.622 (0.753) & \textbf{3.992} (0.086) & 3.885 (0.445) & 1.619 (0.754) & 2.945 (0.761) & \textbf{2.947} (0.676) \\
        Claude Sonnet 4 & 2.476 (0.773) & 3.987 (0.150) & 3.897 (0.410) & \textbf{1.689} (0.664) & 2.920 (0.756) & 2.794 (0.693) \\
        \bottomrule
    \end{tabular}
    \caption{Mean (and standard deviation) of GPT-4.1 evaluations of Oogiri generation.}
    \label{tab:GPT_reordered}
\end{table*}

\section{RQ1: Can LLMs generate humorous responses that humans judge as funny?}

\paragraph{Results}
Human evaluation data (Table \ref{tab:Human_reordered}) reveals several key findings regarding LLM generation abilities. 
Firstly, current models generate Oogiri responses at a level between that of low- and mid-tier humans. 
For instance, the best-performing model, Gemini 2.5 Pro, averaged 1.803 in Overall Funniness, placing it below the human Mid-tier (1.913) but above the Low-tier (1.232). 
Secondly, the primary weakness of LLMs compared to high-tier human responses is a notable deficit in Empathy.
As shown in the radar chart (Figure \ref{fig:Radar_Chart}), the largest performance gap is on this dimension. 
Finally, among the LLMs, Gemini 2.5 Pro was the most proficient, achieving the highest scores across all six dimensions, largely due to its relative strength in Empathy and Clarity.

\paragraph{Discussions}
The results suggest that while LLMs can construct logically coherent and relevant responses, they struggle to create truly compelling humor due to a gap between pattern recognition and genuine human-like cognition. 
The notable deficit in Empathy points to the core of this challenge. 
As revealed by the heatmap of human evaluation dimensions (lower triangle of Figure \ref{fig:dimension_Heat_Human_and_GPT}), Empathy is the dimension most strongly correlated with Overall Funniness for humans. 
This indicates that the ability to resonate with shared human social contexts, cultural norms, and emotions is a foundational element of high-quality humor, and it is precisely this area where current LLMs fall short (Table \ref{tab:example_RQ1}). 
This finding can be effectively interpreted through the lens of incongruity-resolution theories of humor \cite{Suls1972}. 
Our results suggest that LLMs are relatively capable of the first stage of this process: creating incongruity by generating responses with high Novelty while maintaining some Relevance to the topic. 
However, they fail at the crucial second stage---the resolution. 
This is because resolving the incongruity satisfyingly requires Empathy, the ability to ground the unexpected punchline in a shared, relatable human context. 
The LLMs' notable deficit in Empathy prevents them from achieving this humorous resolution, even when their responses are novel and relevant.
Therefore, to generate responses that humans consistently find funny, future development must focus not only on creativity (Novelty) but critically on improving the models' capacity for empathetic reasoning.

\section{RQ2: Can LLMs judge humorous responses that humans judge as funny?}

\begin{figure}[t]
    \setlength\abovecaptionskip{2truemm}
    \centering
    \includegraphics[width=0.9\linewidth]{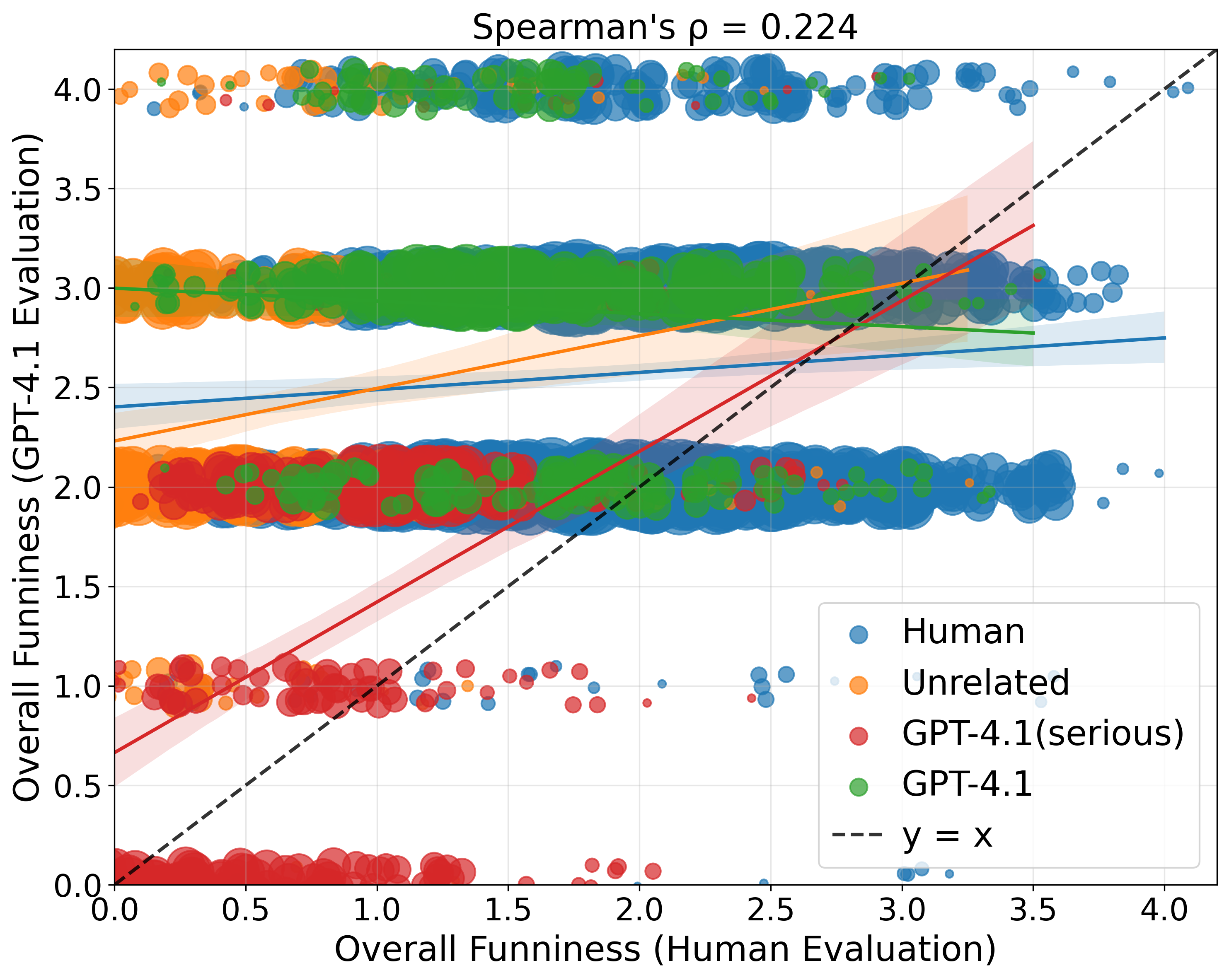}
    \caption{Human vs. LLM ratings: Scatter plot of Overall Funniness. Solid colored lines show the linear regression trend for each response category, with shaded areas indicating the 95 \% confidence interval. The dashed line ($y=x$) represents perfect agreement. GPT-4.1 is a representative example; see Appendix D for results from all models.}
    \label{fig:funniness_scatter}
\end{figure}

\paragraph{Results}
An analysis of the correlation between human and LLM evaluations (Table \ref{tab:spearman_summary}) indicates a limited degree of agreement. 
The overall Spearman correlation for Overall Funniness is low for all models, with Claude Sonnet 4 showing a marginally higher correlation ($\rho=0.266$) than GPT-4.1 ($\rho=0.224$) and Gemini 2.5 Pro ($\rho=0.169$). 
The scatter plots (Figure \ref{fig:funniness_scatter}) visually confirm this weak relationship (Figure \ref{fig:funniness_scatter} shows GPT-4.1; see Appendix D for all models). 


A detailed look at the results of the LLM evaluations for each response type:

\begin{itemize}
    \item For human-generated responses (combining High, Mid, and Low tiers), the LLM evaluations showed a weak positive correlation with human ratings (see Appendix B for details). 
    The accuracy rates for correctly predicting the Overall Funniness order of High, Mid, and Low were 50.9 \% for GPT-4.1, 51.9 \% for Gemini 2.5 Pro, and 54.1 \% for Claude Sonnet 4, respectively, indicating some ability to discern the hierarchy of funniness.
    \item A disagreement between human and LLM evaluators was observed for irrelevant (Unrelated) responses. 
    Humans consistently rated these as not funny, giving them a very low mean Overall Funniness score of 0.681.
    In contrast, LLMs rated these same responses very highly, with mean scores of 2.411 from GPT-4.1 (Table \ref{tab:GPT_reordered}), 3.291 from Gemini 2.5 Pro, and 2.183 from Claude Sonnet 4.
    \item Similarly, for the LLMs' own generated responses, which humans judged to be of lower quality than the Mid-tier, the LLMs rated them at a level comparable to or even higher than the High-tier human responses.
    Furthermore, the LLM evaluations for these responses had a lower standard deviation compared to their evaluations of other response types, indicating less variance in their ratings.
    \item For intentionally non-humorous (serious) responses, the evaluations from humans and LLMs showed a surprisingly high correlation (Figure \ref{fig:funniness_scatter}).
\end{itemize}

\paragraph{Discussions}
The low overall correlation stems from a fundamental divergence in how humans and LLMs evaluate humor. 
The results suggest LLMs have a unique, non-human-like sense of humor. 
This alien ``sensibility'' manifests in clear evaluative biases that differ from human judgment.

One such bias is a positivity bias.
The models systematically overvalued irrelevant (Unrelated) responses.
While humans penalize Unrelated responses for their lack of clarity, relevance, and empathy, LLMs rate them highly.
This bias is also evident elsewhere; while humans differentiate between High, Mid, and Low-tier human responses with an average score gap of over 0.6 points, LLMs evaluate them with only a slight difference in scores (Tables \ref{tab:Human_reordered}, \ref{tab:GPT_reordered}). 
In other words, even for inadequate responses to a topic (such as Unrelated or human Low answers), the models independently judge them as potentially containing clarity, relevance, empathy, and funniness, and consequently assign high scores.
This evaluative leniency could stem from the models' core training objectives. 
Through processes like instruction and preference tuning, they are optimized to be cooperative and to generate positive responses, inadvertently predisposing them to assign favorable scores and hesitate to provide negative feedback.

Another observable bias is a self-preference bias, where LLMs tend to rate responses generated by other LLMs as comparable to high-tier human answers, whereas humans place them below the mid-tier (Tables \ref{tab:Human_reordered}, \ref{tab:GPT_reordered}). 
This self-preference bias aligns with studies by \citet{NEURIPS2024_7f1f0218} and \citet{wataoka2025selfpreferencebiasllmasajudge}.

These biases appear to be symptoms of a deeper difference in evaluation logic.
This is supported by the models' internal evaluation logic, where, in contrast to humans for whom Empathy is the strongest correlate of Funniness (lower triangle in Figure \ref{fig:dimension_Heat_Human_and_GPT}), Novelty is the strongest correlate for LLMs (the upper triangle in Figure \ref{fig:dimension_Heat_Human_and_GPT} shows GPT-4.1 as a representative; see Appendix E for other models).
This tendency to prioritize unexpectedness over contextual and empathetic resonance helps explain why LLMs are prone to the specific evaluative biases mentioned above.

In contrast to this widespread failure in evaluating positive humor, a notable pattern is that agreement is significantly higher for non-humorous content (Figure \ref{fig:funniness_scatter}). 
While LLMs fail to grasp the subjective essence of high-quality humor, they show a strong ability to agree with humans on what is not funny (the serious responses). 
This suggests that ``agreement on the negative'' functions as an easier, objective classification task that does not require the shared subjective sensibility needed for appreciating positive humor.

\section{Conclusion}

In this study, we conducted a multi-dimensional analysis of the humor capabilities of LLMs to both generate and perform evaluations of Oogiri responses. 
This analysis is based on a comprehensive corpus, which we constructed by filtering datasets from prior research and integrating newly collected data, containing responses from both humans and LLMs. 
We then manually annotated this corpus from a six-dimensional perspective.
Regarding generation ability, our results show that LLM performance remains at a level between that of low- and mid-tier humans. 
This performance limitation is primarily attributed to a notable deficit in Empathy.
To enhance humor generation ability, future training should focus on improving Empathy.
Regarding evaluation ability, we found low agreement between human and LLM assessments, revealing a fundamental divergence in evaluation criteria: humans prioritize Empathy, whereas LLMs prioritize Novelty.
This divergence in criteria suggests that humans and LLMs may possess inherently different sensibilities for evaluating humor.


\newpage
\section{Acknowledgments}
This work is partly supported by JST PRESTO Grant Number JPMJPR2366, and also by JSPS KAKENHI Grant Number 22H03651 and 25K03178.

\bibliography{aaai2026}

\clearpage
\onecolumn 
\appendix

\section{A. Prompt template}
\begin{figure}[H]
    \setlength\abovecaptionskip{2truemm}
    \centering
    \includegraphics[width=0.9\linewidth]{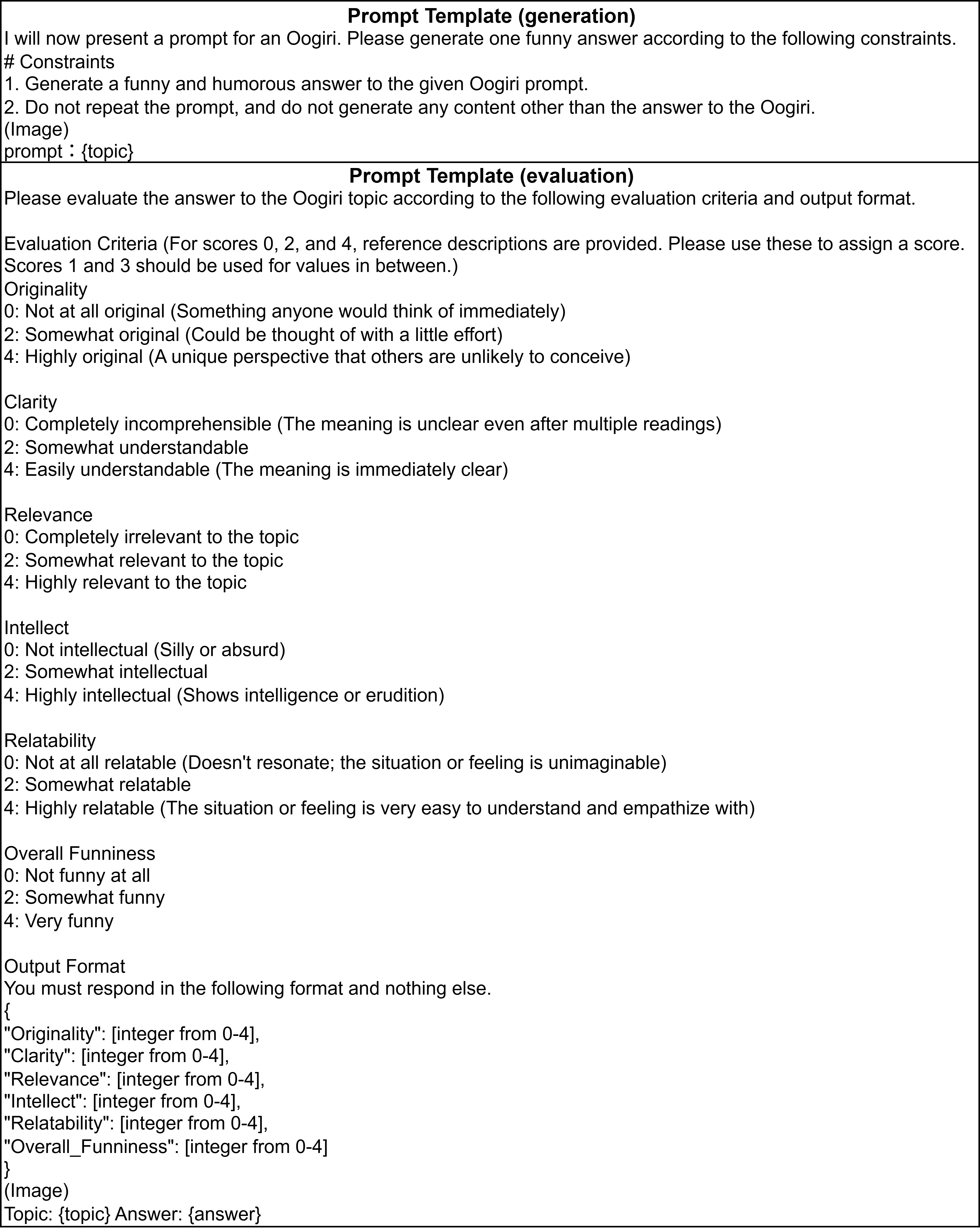}
    \caption{Prompt template of Oogiri generation and evaluation. The original prompt was in Japanese and has been translated here for clarity.}
    \label{fig:prompt_template}
\end{figure}

Figure \ref{fig:prompt_template} is the example of a prompt template for generation and evaluation.
Please note that the template shown has been translated from the original Japanese into English for publication purposes.

\section{B. Analysis of LLM Evaluations on Human Response Tiers}
\begin{figure}[H]
    \setlength\abovecaptionskip{2truemm}
    \centering
    \includegraphics[width=1\linewidth]{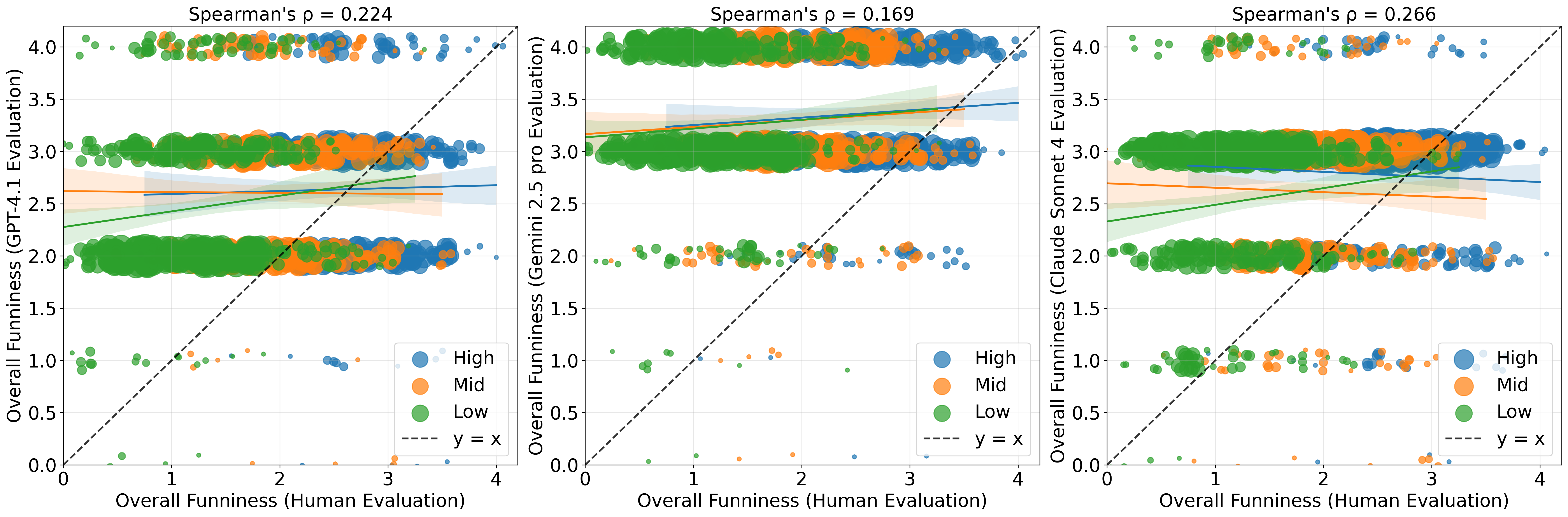}
    \caption{Scatter plot of models evaluations for human responses. Solid colored lines show the linear regression trend for each response category, with shaded areas indicating the 95 \% confidence interval. The dashed line ($y = x$) represents perfect agreement.}
    \label{fig:scatter_human_res}
\end{figure}

Figure \ref{fig:scatter_human_res} illustrates the scatter plots of human vs LLM evaluations, specifically for the human-rated High, Mid, and Low-tier responses. 
As mentioned in the RQ2 section of the main text, these plots show that while the human-rated tiers form a clear gradation, the LLMs tend to assign high scores more generally across these tiers. 
This visually confirms that the LLMs' ability to correctly rank the responses by quality is limited.

\section{C. Evaluations results of other models}
Table \ref{tab:Gemini_reordered} and Table \ref{tab:Claude_reordered} show the evaluation results from Gemini 2.5 Pro and Claude Sonnet 4, respectively.
According to Gemini 2.5 Pro's own evaluations, the human High-tier responses received the highest average Overall Funniness score (3.363). However, this was nearly equivalent to the score it assigned to its own generated responses (3.346) , confirming the cross-model bias where LLMs tend to rate responses from other LLMs highly. 
On the other hand, when comparing how each model evaluated the AI-generated responses, all three models (GPT-4.1, Gemini 2.5 Pro, and Claude Sonnet 4) rated Gemini 2.5 Pro's responses as the most humorous. This finding aligns with the human evaluations, where Gemini 2.5 Pro was also judged to be the most proficient generator. 

\begin{table}[H]
    \centering
    \begin{tabular}{l|rrrrrr}
        \toprule
        & {\textbf{Novelty}} & {\textbf{Clarity}} & {\textbf{Relevance}} & {\textbf{Intelligence}} & {\textbf{Empathy}} & {\textbf{Overall Funniness}} \\
        \midrule
        High & 2.679 (0.989) & 3.940 (0.342) & 3.932 (0.435) & 1.190 (1.129) & 2.664 (1.185) & \textbf{3.363} (0.702) \\
        Mid & 2.684 (0.970) & 3.912 (0.368) & 3.937 (0.373) & 1.276 (1.221) & 2.659 (1.167) & 3.296 (0.707) \\
        Low & 2.764 (0.995) & 3.762 (0.706) & 3.872 (0.502) & 1.313 (1.236) & 2.313 (1.236) & 3.238 (0.747) \\
        Unrelated & \textbf{3.243} (0.798) & 3.266 (1.087) & 3.201 (1.240) & \textbf{1.596} (1.186) & 2.185 (1.358) & 3.291 (0.933) \\
        \midrule
        GPT-4.1 (serious) & 1.043 (1.254) & 3.980 (0.255) & 3.955 (0.337) & 0.779 (1.062) & 2.218 (1.451) & 1.333 (1.424) \\
        GPT-4.1 & 2.564 (0.871) & 3.977 (0.165) & 3.952 (0.301) & 0.957 (1.075) & 2.792 (1.061) & 3.203 (0.590) \\
        Gemini 2.5 Pro & 2.602 (0.873) & \textbf{3.987} (0.132) & \textbf{3.987} (0.132) & 0.967 (1.094) & \textbf{2.880} (1.066) & 3.346 (0.589) \\
        Claude Sonnet 4 & 2.459 (0.884) & 3.980 (0.186) & 3.947 (0.332) & 1.048 (1.066) & 2.777 (1.111) & 3.155 (0.666) \\
        \bottomrule
    \end{tabular}
    \caption{Mean (and standard deviation) of Gemini 2.5 Pro evaluations of Oogiri generation.}
    \label{tab:Gemini_reordered}
\end{table}

\begin{table}[H]
    \centering
    \begin{tabular}{l|rrrrrr}
        \toprule
        & {\textbf{Novelty}} & {\textbf{Clarity}} & {\textbf{Relevance}} & {\textbf{Intelligence}} & {\textbf{Empathy}} & {\textbf{Overall Funniness}} \\
        \midrule
        High & 2.747 (0.763) & 3.815 (0.627) & 3.749 (0.700) & \textbf{1.870} (0.742) & 2.840 (0.701) & 2.777 (0.682) \\
        Mid & 2.639 (0.860) & 3.649 (0.867) & 3.609 (0.926) & 1.822 (0.845) & 2.702 (0.835) & 2.614 (0.815) \\
        Low & 2.632 (0.875) & 3.481 (1.017) & 3.506 (1.029) & 1.712 (0.820) & 2.571 (0.876) & 2.526 (0.850) \\
        Unrelated & 2.699 (1.005) & 2.669 (1.434) & 2.464 (1.556) & 1.657 (0.894) & 2.025 (1.173) & 2.183 (1.077) \\
        \midrule
        GPT-4.1 (serious) & 1.263 (1.211) & 3.950 (0.378) & 3.378 (1.186) & 1.216 (0.961) & 2.251 (1.090) & 1.331 (1.205) \\
        GPT-4.1 & 2.759 (0.640) & 3.980 (0.199) & 3.922 (0.377) & 1.842 (0.533) & \textbf{3.015} (0.481) & 2.845 (0.502) \\
        Gemini 2.5 Pro & \textbf{2.789} (0.650) & 3.982 (0.218) & 3.905 (0.455) & 1.835 (0.624) & 3.013 (0.533) & \textbf{2.892} (0.517) \\
        Claude Sonnet 4 & 2.719 (0.674) & \textbf{3.987} (0.150) & \textbf{3.965} (0.281) & 1.862 (0.575) & 2.955 (0.543) & 2.830 (0.541) \\
        \bottomrule
    \end{tabular}
    \caption{Mean (and standard deviation) of Claude Sonnet 4 evaluations of Oogiri generation.}
    \label{tab:Claude_reordered}
\end{table}

\section{D. Overall Correlation Between Human and LLM Evaluations}
\begin{figure}[H]
    \setlength\abovecaptionskip{2truemm}
    \centering
    \includegraphics[width=0.85\linewidth]{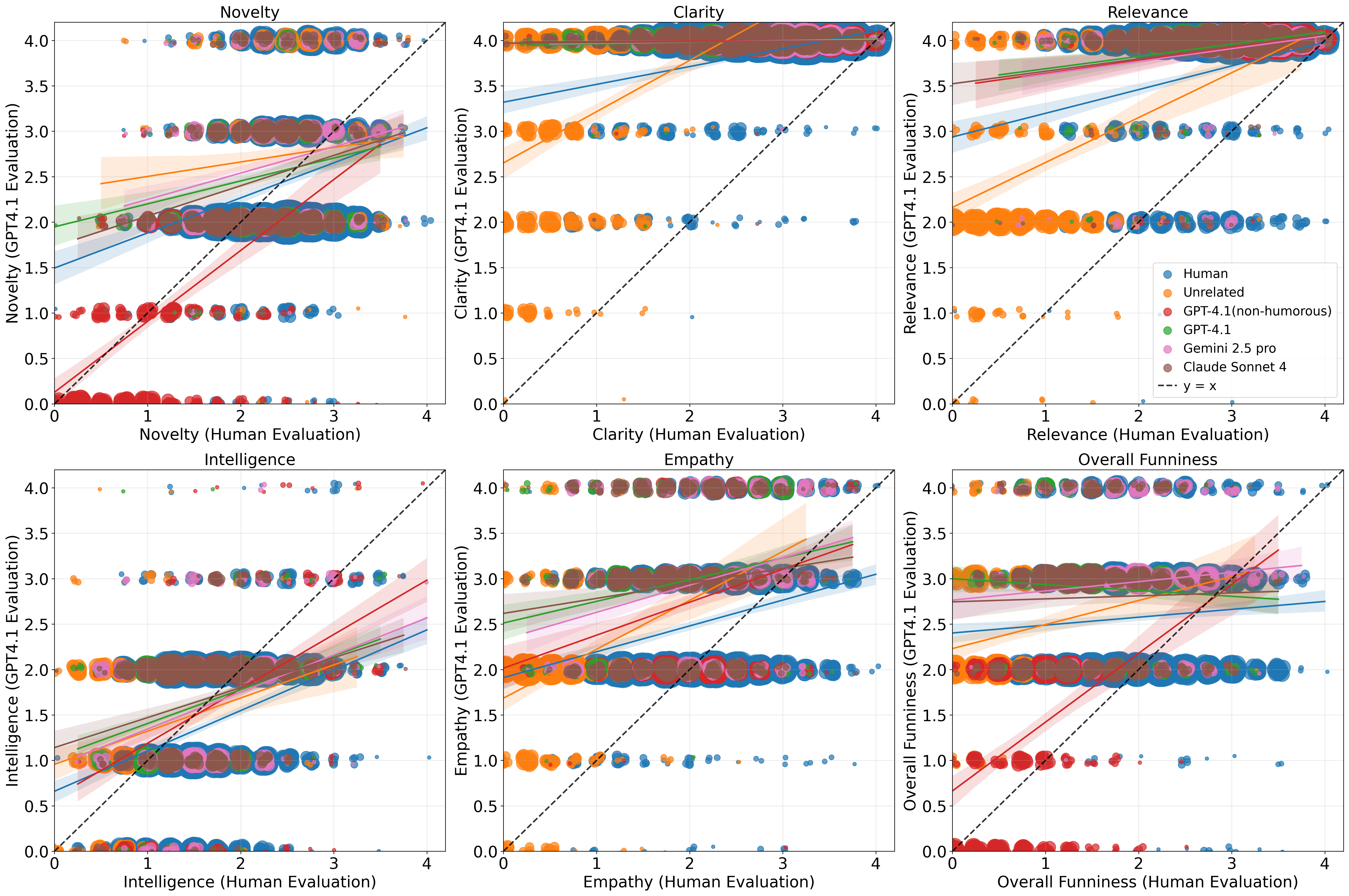}
    \caption{Scatter plots (Human vs GPT-4.1)}
    \label{fig:GPT_scatter_all}
\end{figure}

\begin{figure}[H]
    \setlength\abovecaptionskip{2truemm}
    \centering
    \includegraphics[width=0.85\linewidth]{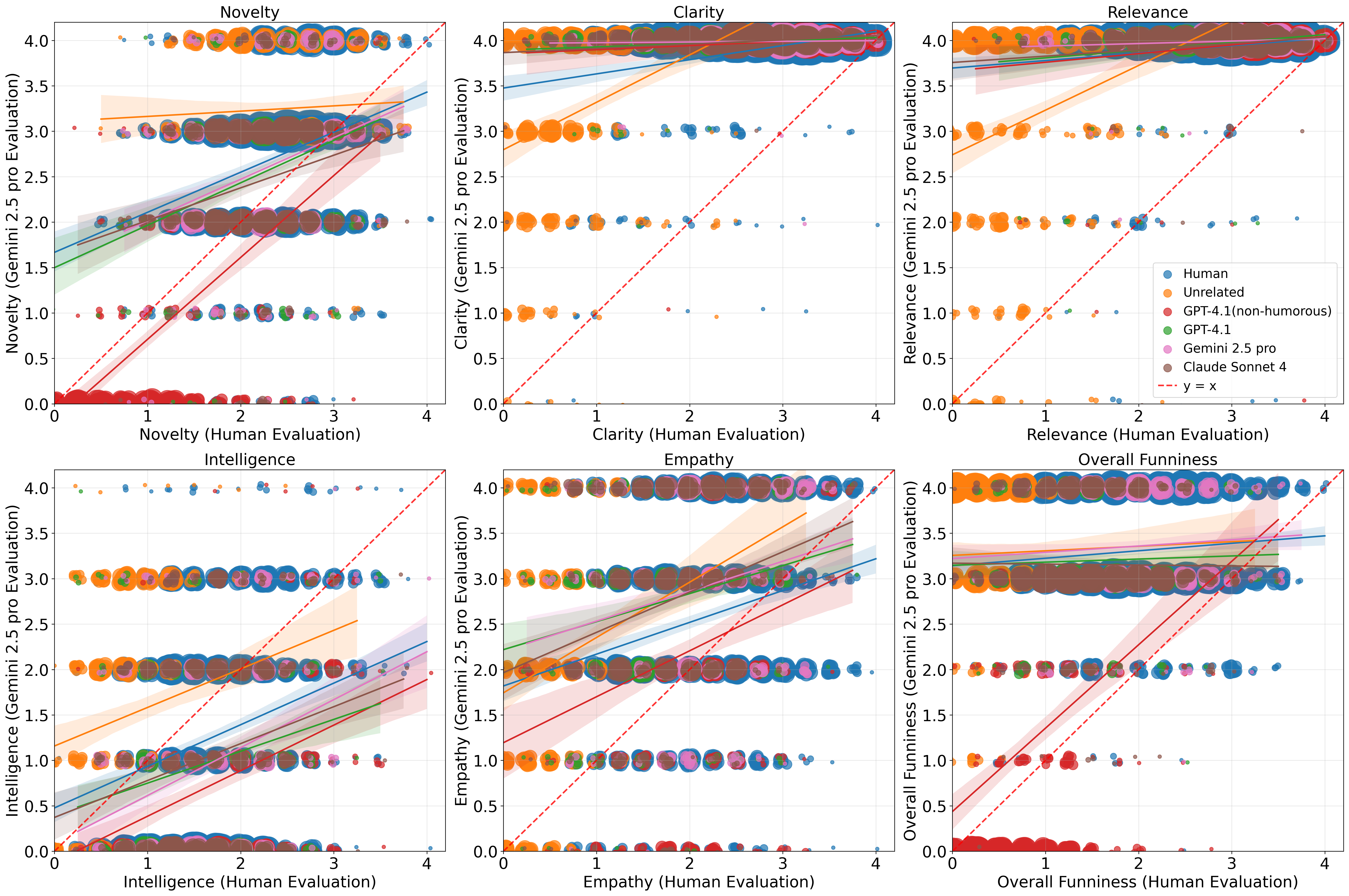}
    \caption{Scatter plots (Human vs Gemini 2.5 Pro)}
    \label{fig:Gemini_scatter_all}
\end{figure}

\begin{figure}[H]
    \setlength\abovecaptionskip{2truemm}
    \centering
    \includegraphics[width=0.85\linewidth]{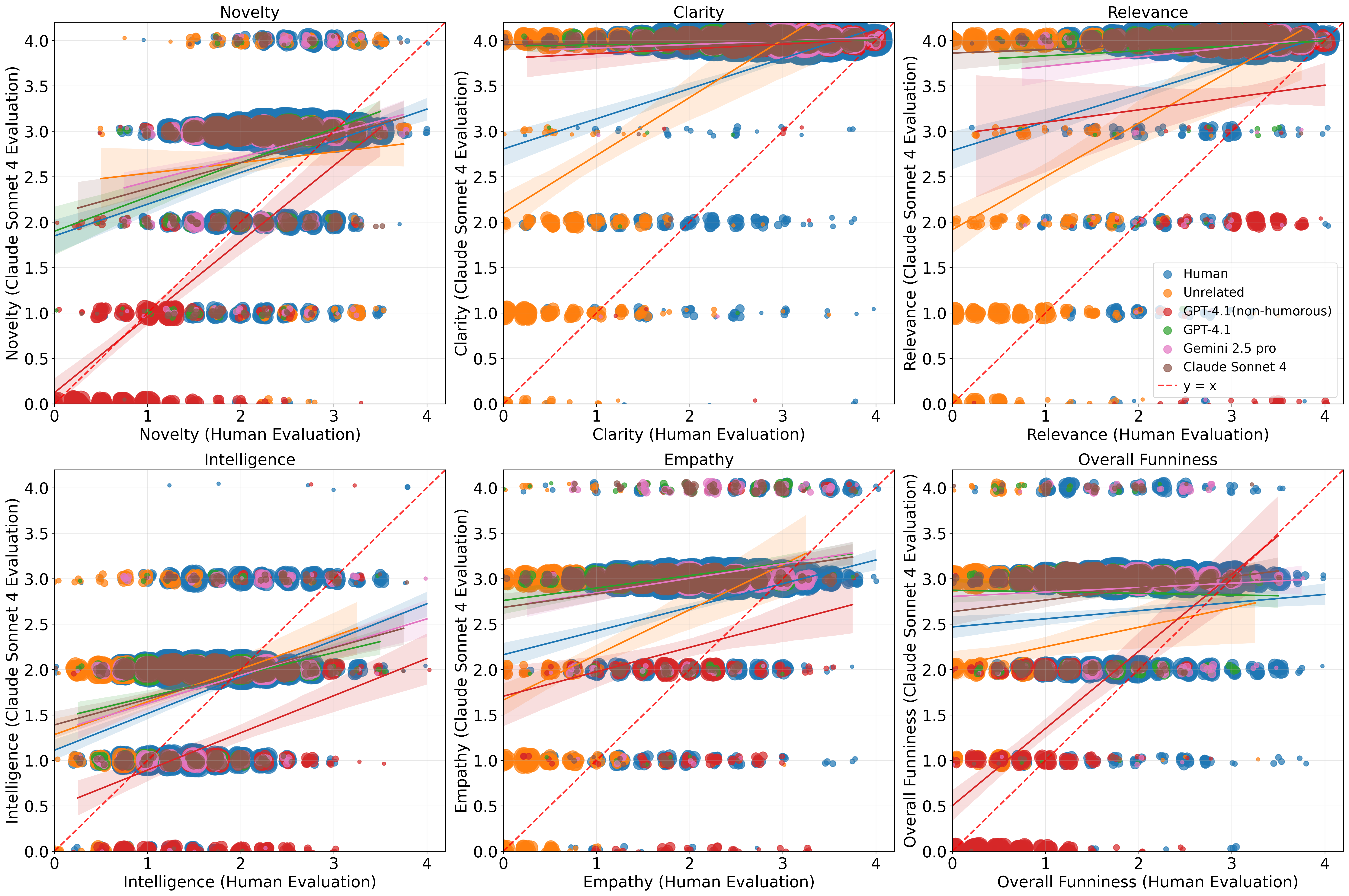}
    \caption{Scatter plots (Human vs Claude Sonnet 4)}
    \label{fig:Claude_scatter_all}
\end{figure}

Figures \ref{fig:GPT_scatter_all}, \ref{fig:Gemini_scatter_all}, and \ref{fig:Claude_scatter_all} show the scatter plots comparing human evaluations with those of each model, while Tables \ref{tab:gpt_correlation}, \ref{tab:gemini_correlation}, and \ref{tab:claude_correlation} present the results for the Pearson and Spearman correlation coefficients, both overall and for each axis. From the scatter plots, it is visually apparent that all models tend to assign particularly high scores for Clarity and Relevance. Furthermore, the overall correlation coefficients for the individual dimensions are often higher than the correlation for Overall Funniness. This indicates that a high correlation on more objective dimensions does not necessarily translate to a high correlation on the complex, subjective dimension of Overall Funniness.

\begin{table}[H]
    \centering
    \resizebox{\textwidth}{!}{%
    \begin{tabular}{l|cc|cc|cc|cc|cc|cc}
        \toprule
        & \multicolumn{2}{c|}{\textbf{Novelty}} & \multicolumn{2}{c|}{\textbf{Clarity}} & \multicolumn{2}{c|}{\textbf{Relevance}} & \multicolumn{2}{c|}{\textbf{Intelligence}} & \multicolumn{2}{c|}{\textbf{Empathy}} & \multicolumn{2}{c}{\textbf{Overall Funniness}} \\
        & Pearson & Spearman & Pearson & Spearman & Pearson & Spearman & Pearson & Spearman & Pearson & Spearman & Pearson & Spearman \\
        \midrule
        \textbf{Overall} & 0.429* & 0.361* & 0.426* & 0.379* & 0.503* & 0.441* & 0.362* & 0.352* & 0.320* & 0.303* & 0.242* & 0.224* \\
        \midrule
        High & 0.224* & 0.213* & 0.197* & 0.193* & 0.212* & 0.212* & 0.412* & 0.405* & 0.210* & 0.205* & 0.023 & 0.030 \\
        Mid & 0.298* & 0.291* & 0.311* & 0.229* & 0.211* & 0.206* & 0.363* & 0.341* & 0.206* & 0.211* & -0.007 & 0.026 \\
        Low & 0.340* & 0.319* & 0.383* & 0.399* & 0.339* & 0.329* & 0.360* & 0.341* & 0.282* & 0.267* & 0.121* & 0.127* \\
        Unrelated & 0.104* & 0.120* & 0.370* & 0.383* & 0.337* & 0.323* & 0.256* & 0.242* & 0.335* & 0.329* & 0.181* & 0.175* \\
        GPT-4.1 (serious) & 0.579* & 0.577* & 0.109* & 0.083 & 0.250* & 0.228* & 0.472* & 0.478* & 0.299* & 0.291* & 0.440* & 0.450* \\
        GPT-4.1 & 0.215* & 0.212* & 0.117* & 0.128* & 0.269* & 0.254* & 0.362* & 0.372* & 0.248* & 0.248* & -0.071 & -0.074 \\
        Gemini 2.5 Pro & 0.237* & 0.236* & 0.115* & 0.085 & 0.213* & 0.198* & 0.345* & 0.338* & 0.275* & 0.264* & 0.105* & 0.084 \\
        Claude Sonnet 4 & 0.258* & 0.223* & 0.041 & 0.021 & 0.238* & 0.211* & 0.316* & 0.292* & 0.154* & 0.160* & 0.032 & 0.026 \\
        \bottomrule
    \end{tabular}%
    }
    \caption{Correlation coefficients between human evaluation and GPT-4.1 evaluation. In addition to calculating the correlation coefficients, we performed statistical significance testing for each value. The p-value associated with each Pearson and Spearman coefficient is derived from a t-test against the null hypothesis that the true correlation is zero. Significance is reported in the tables using an asterisk ($p < 0.05$).}
    \label{tab:gpt_correlation}
\end{table}

\begin{table}[H]
    \centering
    \resizebox{\textwidth}{!}{%
    \begin{tabular}{l|cc|cc|cc|cc|cc|cc}
        \toprule
        & \multicolumn{2}{c|}{\textbf{Novelty}} & \multicolumn{2}{c|}{\textbf{Clarity}} & \multicolumn{2}{c|}{\textbf{Relevance}} & \multicolumn{2}{c|}{\textbf{Intelligence}} & \multicolumn{2}{c|}{\textbf{Empathy}} & \multicolumn{2}{c}{\textbf{Overall Funniness}} \\
        & Pearson & Spearman & Pearson & Spearman & Pearson & Spearman & Pearson & Spearman & Pearson & Spearman & Pearson & Spearman \\
        \midrule
        \textbf{Overall} & 0.474* & 0.393* & 0.386* & 0.356* & 0.374* & 0.328* & 0.206* & 0.188* & 0.255* & 0.248* & 0.242* & 0.169* \\
        \midrule
        High & 0.246* & 0.224* & 0.153* & 0.169* & 0.089 & 0.108* & 0.373* & 0.355* & 0.220* & 0.191* & 0.065 & 0.058 \\
        Mid & 0.334* & 0.339* & 0.214* & 0.204* & 0.114* & 0.111* & 0.290* & 0.264* & 0.197* & 0.196* & 0.065 & 0.060 \\
        Low & 0.376* & 0.352* & 0.307* & 0.308* & 0.199* & 0.203* & 0.290* & 0.253* & 0.220* & 0.209* & 0.074 & 0.036 \\
        Unrelated & 0.046 & 0.018 & 0.337* & 0.389* & 0.288* & 0.299* & 0.219* & 0.210* & 0.276* & 0.277* & 0.032 & -0.035 \\
        GPT-4.1 (serious) & 0.593* & 0.591* & 0.128* & 0.086 & 0.176* & 0.153* & 0.350* & 0.352* & 0.247* & 0.248* & 0.422* & 0.433* \\
        GPT-4.1 & 0.325* & 0.299* & 0.176* & 0.179* & 0.196* & 0.160* & 0.214* & 0.215* & 0.224* & 0.231* & 0.040 & 0.035 \\
        Gemini 2.5 Pro & 0.323* & 0.318* & 0.041 & 0.050 & 0.138* & 0.134* & 0.308* & 0.278* & 0.216* & 0.211* & 0.082 & 0.087 \\
        Claude Sonnet 4 & 0.242* & 0.218* & 0.204* & 0.134* & 0.149* & 0.127* & 0.242* & 0.221* & 0.282* & 0.269* & -0.10 & -0.027 \\
        \bottomrule
    \end{tabular}%
    }
    \caption{Correlation coefficients between human evaluation and Gemini 2.5 Pro evaluation.}
    \label{tab:gemini_correlation}
\end{table}

\begin{table}[H]
    \centering
    \resizebox{\textwidth}{!}{%
    \begin{tabular}{l|cc|cc|cc|cc|cc|cc}
        \toprule
        & \multicolumn{2}{c|}{\textbf{Novelty}} & \multicolumn{2}{c|}{\textbf{Clarity}} & \multicolumn{2}{c|}{\textbf{Relevance}} & \multicolumn{2}{c|}{\textbf{Intelligence}} & \multicolumn{2}{c|}{\textbf{Empathy}} & \multicolumn{2}{c}{\textbf{Overall Funniness}} \\
        & Pearson & Spearman & Pearson & Spearman & Pearson & Spearman & Pearson & Spearman & Pearson & Spearman & Pearson & Spearman \\
        \midrule
        \textbf{Overall} & 0.454* & 0.375* & 0.431* & 0.383* & 0.384* & 0.315* & 0.290* & 0.275* & 0.300* & 0.278* & 0.293* & 0.266* \\
        \midrule
        High & 0.188* & 0.168* & 0.276* & 0.246* & 0.238* & 0.220* & 0.395* & 0.389* & 0.263* & 0.253* & -0.046 & -0.039 \\
        Mid & 0.279* & 0.287* & 0.307* & 0.275* & 0.246* & 0.228* & 0.368* & 0.334* & 0.173* & 0.172* & -0.035 & -0.022 \\
        Low & 0.327* & 0.316* & 0.352* & 0.354* & 0.301* & 0.307* & 0.314* & 0.305* & 0.270* & 0.249* & 0.124* & 0.106* \\
        Unrelated & 0.074 & 0.064 & 0.312* & 0.335* & 0.274* & 0.282* & 0.247* & 0.227* & 0.260* & 0.281* & 0.117* & 0.117* \\
        GPT-4.1 (serious) & 0.568* & 0.567* & 0.100* & 0.078 & 0.071 & 0.062 & 0.315* & 0.298* & 0.175* & 0.179* & 0.460* & 0.479* \\
        GPT-4.1 & 0.358* & 0.327* & 0.075 & 0.098 & 0.100* & 0.095 & 0.300* & 0.300* & 0.220* & 0.223* & -0.024 & -0.034 \\
        Gemini 2.5 Pro & 0.254* & 0.265* & 0.130* & 0.122* & 0.153* & 0.142* & 0.317* & 0.298* & 0.210* & 0.208* & 0.066 & 0.061 \\
        Claude Sonnet 4 & 0.253* & 0.183* & 0.077 & 0.099* & 0.097 & 0.052 & 0.314* & 0.277* & 0.193* & 0.197* & 0.158* & 0.147* \\
        \bottomrule
    \end{tabular}%
    }
    \caption{Correlation coefficients between human evaluation and Claude Sonnet 4 evaluation.}
    \label{tab:claude_correlation}
\end{table}

\newpage
\section{E. Correlation of Evaluation Dimensions for Gemini 2.5 Pro and Claude Sonnet 4}
\begin{figure}[H]
    \setlength\abovecaptionskip{2truemm}
    \centering
    \includegraphics[width=0.8\linewidth]{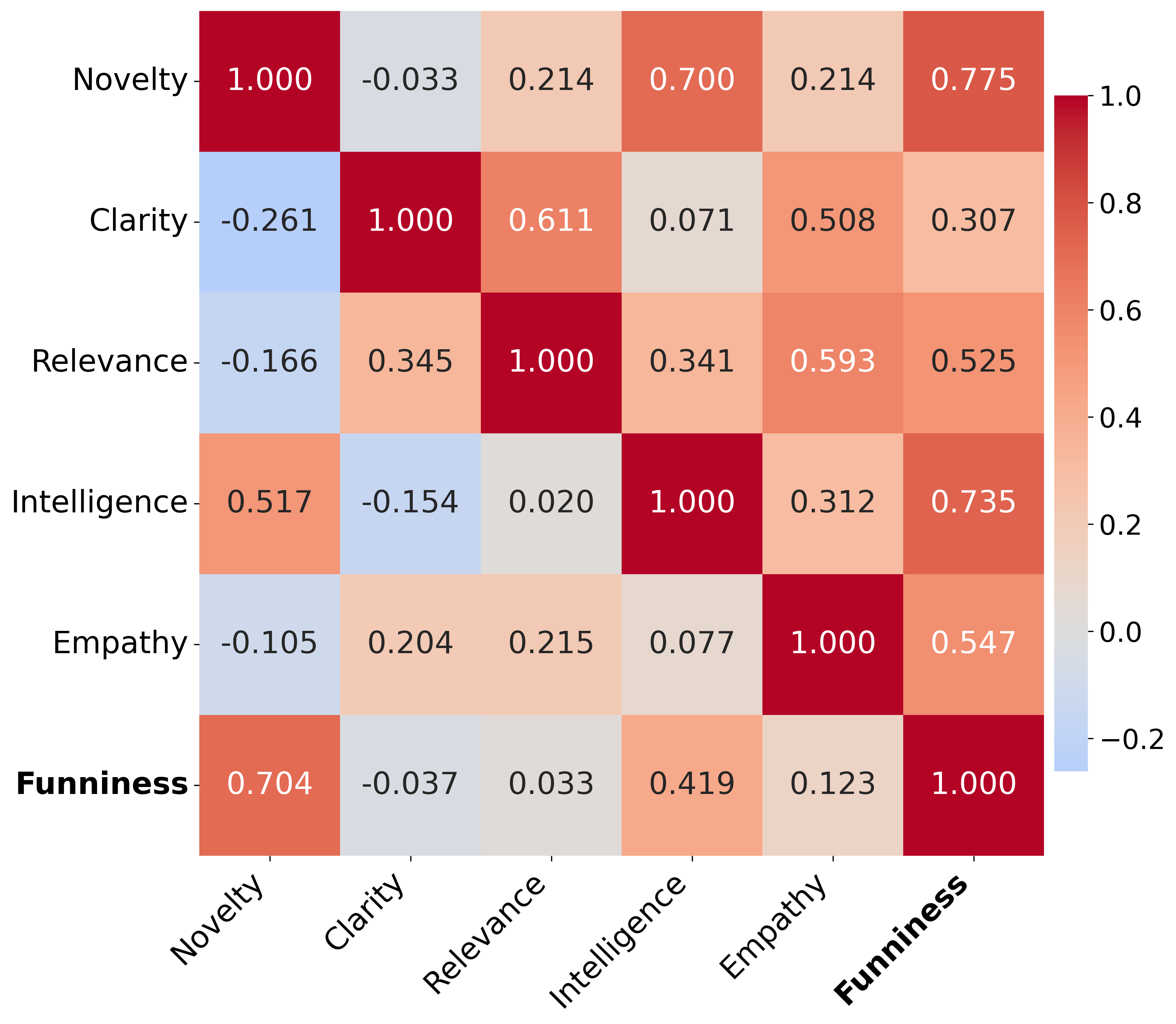}
    \caption{Spearman correlation matrix of Gemini 2.5 Pro and Claude Sonnet 4 evaluation. Gemini 2.5 Pro evaluations are in the lower triangle, and Claude Sonnet 4 evaluations are in the upper triangle.}
    \label{fig:dimension_Heat_Gemini_Claude}
\end{figure}

The lower triangle of Figure \ref{fig:dimension_Heat_Gemini_Claude} shows the heatmap of the correlations between evaluation dimensions for Gemini 2.5 Pro, while the upper triangle shows the same for Claude Sonnet 4. 
Similar to GPT-4.1, the results for both models show that 
Novelty has the highest correlation with Overall Funniness. 
As a distinct characteristic, Claude Sonnet 4 also showed a high correlation between  
Intelligence and Overall Funniness, a pattern not seen in the other models. 

\section{F. Histograms of Human and LLMs evaluations}
\begin{figure}[H]
    \setlength\abovecaptionskip{2truemm}
    \centering
    \includegraphics[width=1\linewidth]{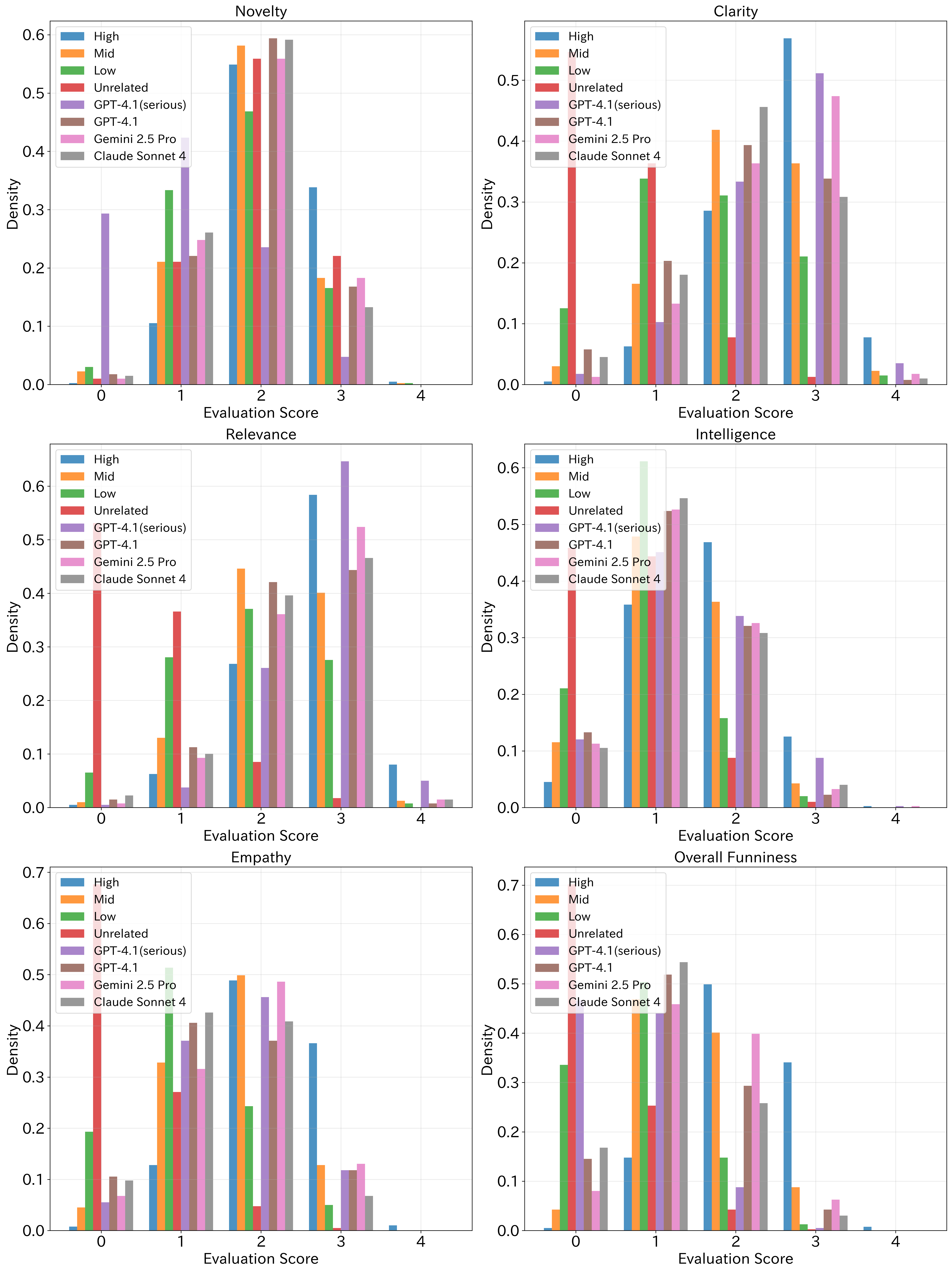}
    \caption{Human evaluations histograms.}
    \label{fig:Hist_Human}
\end{figure}

\begin{figure}[H]
    \setlength\abovecaptionskip{2truemm}
    \centering
    \includegraphics[width=1\linewidth]{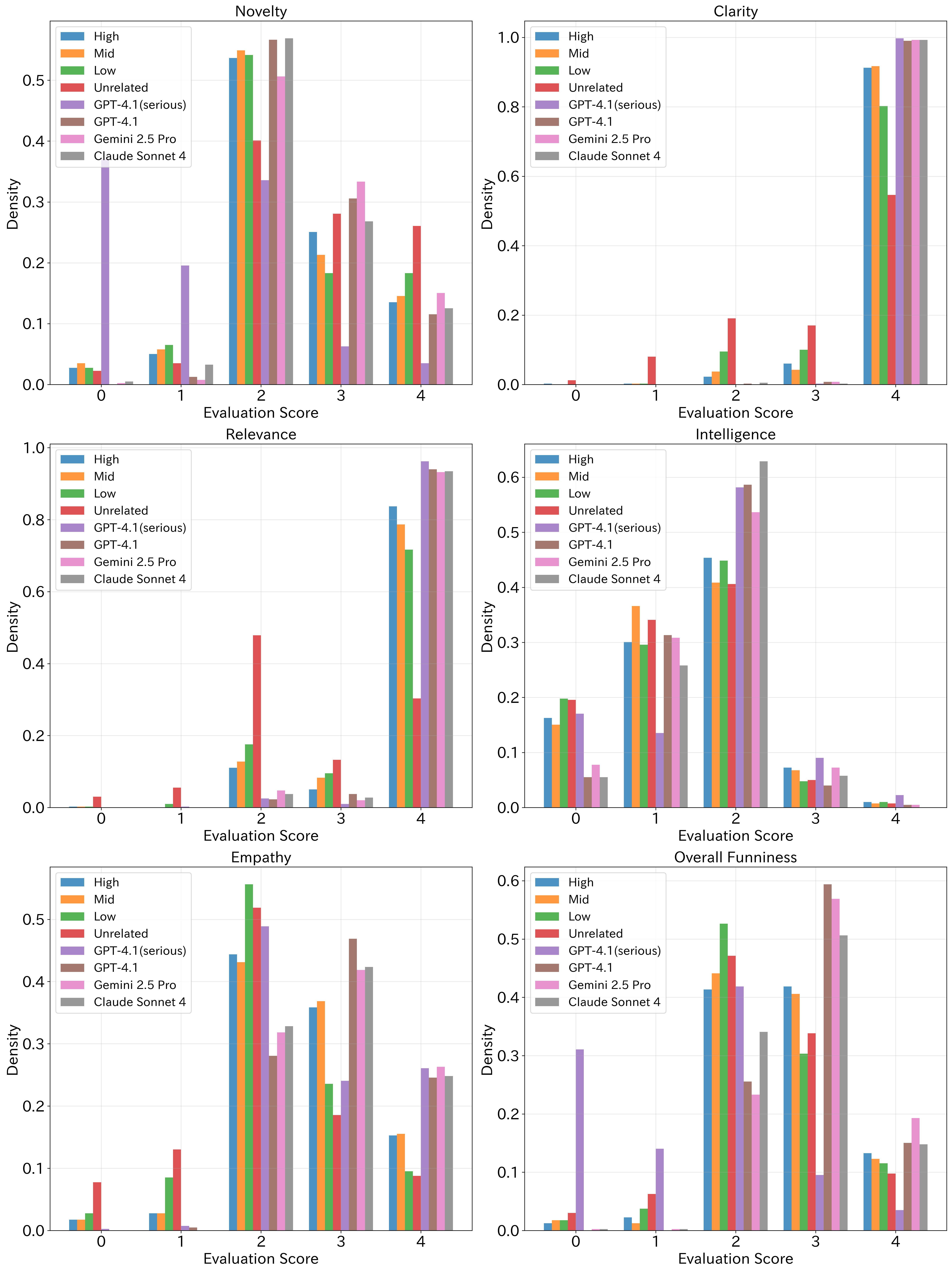}
    \caption{GPT-4.1 evaluations histograms.}
    \label{fig:Hist_GPT}
\end{figure}

\begin{figure}[H]
    \setlength\abovecaptionskip{2truemm}
    \centering
    \includegraphics[width=1\linewidth]{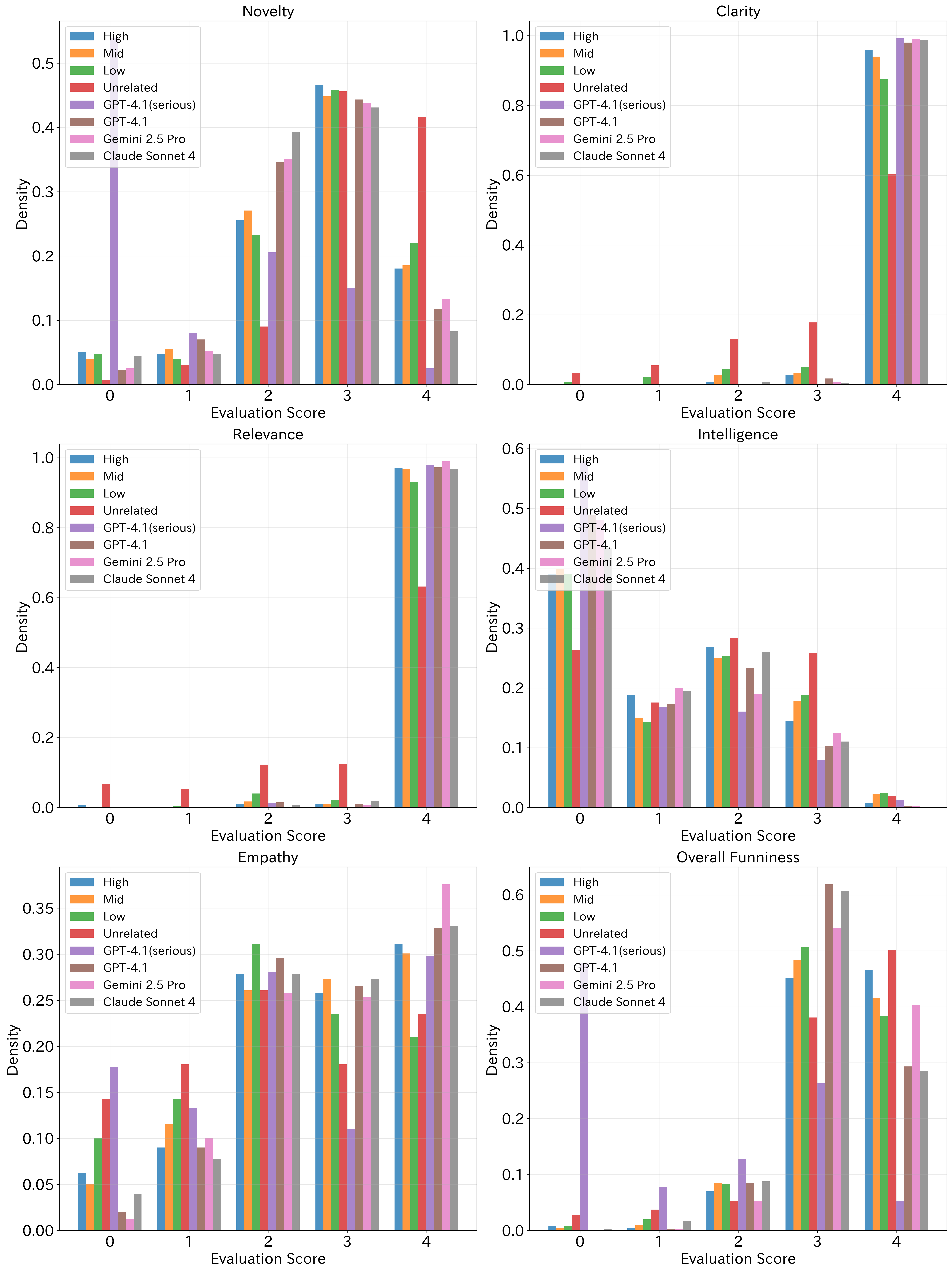}
    \caption{Gemini 2.5 Pro evaluations histograms.}
    \label{fig:Hist_Gemini}
\end{figure}

\begin{figure}[H]
    \setlength\abovecaptionskip{2truemm}
    \centering
    \includegraphics[width=1\linewidth]{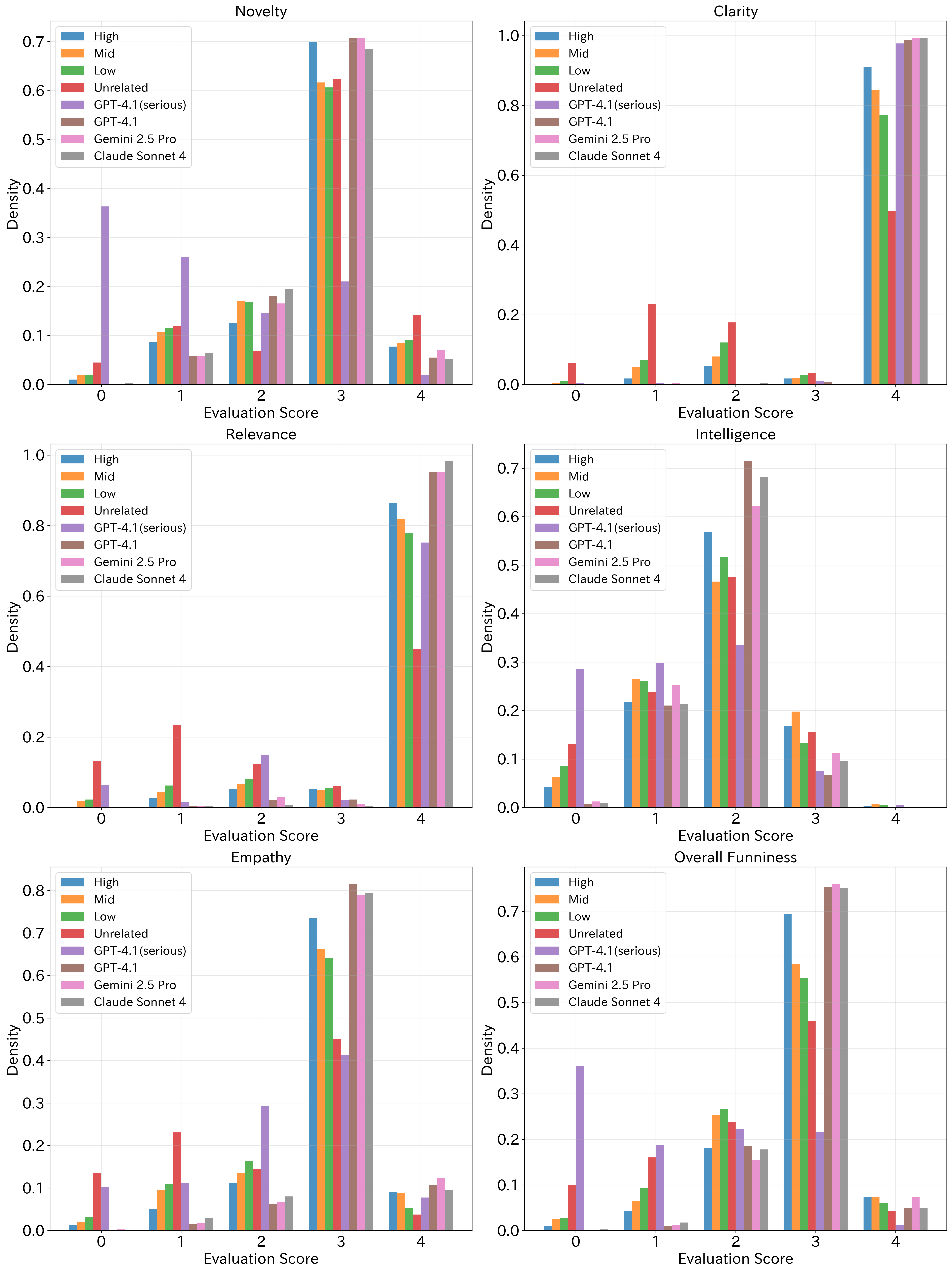}
    \caption{Claude Sonnet 4 evaluations histograms.}
    \label{fig:Hist_Claude}
\end{figure}

Figures \ref{fig:Hist_Human}, \ref{fig:Hist_GPT}, \ref{fig:Hist_Gemini} and \ref{fig:Hist_Claude} are the histograms of human and LLM evaluations.

\section{G. Difference between text topics and image topics}
To analyze how the prompt format (text-based vs. image-based) affects humor generation and evaluation, we aggregated the results separately for each prompt type.

\paragraph{Human Evaluation Tendencies (Tables \ref{tab:Human_text} and \ref{tab:Human_image})}
Human evaluations reveal that the prompt modality has a clear impact on funniness scores for both human and LLM-generated responses. For human-generated responses across all tiers (High, Mid, and Low), image-based topics consistently received higher Overall Funniness scores than text-based topics (Tables \ref{tab:Human_text} and \ref{tab:Human_image}). 
This suggests that image topics, perhaps by providing richer context, may be a more fertile ground for generating responses that humans perceive as funny.

While the responses generated by most LLMs also received higher scores on image topics, their performance gain was not uniform. 
For instance, the score for the best-performing model, Gemini 2.5 Pro, increased from 1.719 on text topics to 1.887 on image topics. 
Claude Sonnet 4, however, was a notable exception, with its score decreasing from 1.552 on text topics to 1.456 on image topics. 
Furthermore, the performance increase for Gemini (0.168 points) is smaller than the increase observed for High-tier human responses (0.207 points). 
This indicates that while LLMs can benefit from the format of image topics, their ability to capitalize on the rich, creative potential provided by visual context is inconsistent and generally less adept than that of skilled humans.

\paragraph{LLM Evaluation Tendencies (Tables \ref{tab:GPT_text}--\ref{tab:Claude_image})}
In the LLM evaluations, the prompt format revealed interesting results, with each model's ``sensibilities'' and biases manifesting differently.

\begin{itemize}
    \item GPT-4.1 Evaluation Tendency: GPT-4.1 tended to rate the Clarity of AI-generated responses for image topics as nearly a perfect 4.000 (Table \ref{tab:GPT_image}). This suggests a potential bias of overvaluing the simplicity and clarity of short text captions for images.
    \item Gemini 2.5 Pro Evaluation Tendency: Gemini 2.5 Pro's evaluation bias was particularly pronounced on text topics. It rated Unrelated responses as funnier (3.320) than even the High-tier human responses (3.315) (Table \ref{tab:Gemini_text}). This shows that its tendency for ``context-negligence''---ignoring the prompt and prioritizing standalone novelty---is even stronger in the less constrained domain of text.
    \item Claude Sonnet 4 Evaluation Tendency: Claude Sonnet 4 exhibited a unique characteristic, in contrast to the other two models and to humans: it consistently rated text-based humor more highly. For human High-tier responses, it scored text topics (2.945) significantly higher than image topics (2.608) (Tables \ref{tab:Claude_text} and \ref{tab:Claude_image}). This pattern suggests Claude Sonnet 4 may have a ``preference'' for text-based humor.
\end{itemize}

From this analysis, it is clear that the prompt modality (text vs. image) is not merely a difference in input format but an important factor that affects both the generation performance of LLMs (as rated by humans) and their own evaluation logic. The different ways biases manifest and the existence of model-specific ``preferences'' for a certain modality further highlight the complexity of the LLMs' ``alien sensibility.''

\begin{table}[H]
    \centering
    \begin{tabular}{l|rrrrrr}
        \toprule
        & {\textbf{Novelty}} & {\textbf{Clarity}} & {\textbf{Relevance}} & {\textbf{Intelligence}} & {\textbf{Empathy}} & {\textbf{Overall Funniness}} \\
        \midrule
        High & \textbf{2.594} (0.555) & \textbf{2.974} (0.730) & \textbf{3.106} (0.646) & \textbf{2.205} (0.702) & \textbf{2.606} (0.664) & \textbf{2.460} (0.617) \\
        Mid & 2.301 (0.670) & 2.525 (0.807) & 2.741 (0.656) & 1.822 (0.725) & 2.096 (0.711) & 1.821 (0.629) \\
        Low & 2.082 (0.685) & 2.016 (1.007) & 2.381 (0.921) & 1.394 (0.628) & 1.481 (0.788) & 1.116 (0.627) \\
        Unrelated & 2.255 (0.657) & 0.750 (0.671) & 0.895 (0.727) & 0.991 (0.596) & 0.621 (0.595) & 0.546 (0.510) \\
        \midrule
        GPT-4.1 (serious) & 1.636 (0.797) & 2.712 (0.782) & 2.960 (0.642) & 1.918 (0.728) & 1.965 (0.743) & 1.084 (0.646) \\
        GPT-4.1 & 2.319 (0.599) & 2.390 (0.836) & 2.798 (0.661) & 1.672 (0.679) & 1.842 (0.787) & 1.588 (0.696) \\
        Gemini 2.5 Pro & 2.272 (0.589) & 2.701 (0.758) & 2.930 (0.624) & 1.734 (0.683) & 2.029 (0.716) & 1.719 (0.650) \\
        Claude Sonnet 4 & 2.156 (0.582) & 2.591 (0.757) & 2.938 (0.564) & 1.736 (0.692) & 1.936 (0.683) & 1.552 (0.654) \\
        \bottomrule
    \end{tabular}
    \caption{Mean and standard deviation of human evaluations for each response type (text topics only)}
    \label{tab:Human_text}
\end{table}

\begin{table}[H]
    \centering
    \begin{tabular}{l|rrrrrr}
        \toprule
        & {\textbf{Novelty}} & {\textbf{Clarity}} & {\textbf{Relevance}} & {\textbf{Intelligence}} & {\textbf{Empathy}} & {\textbf{Overall Funniness}} \\
        \midrule
        High & \textbf{2.667} (0.599) & \textbf{3.084} (0.647) & 2.985 (0.681) & \textbf{1.872} (0.676) & \textbf{2.624} (0.649) & \textbf{2.667} (0.663) \\
        Mid & 2.338 (0.568) & 2.618 (0.837) & 2.563 (0.778) & 1.642 (0.596) & 2.103 (0.720) & 2.005 (0.708) \\
        Low & 2.201 (0.691) & 2.020 (0.843) & 2.134 (0.812) & 1.323 (0.599) & 1.539 (0.686) & 1.348 (0.670) \\
        Unrelated & 2.500 (0.577) & 1.046 (0.703) & 0.975 (0.726) & 1.073 (0.627) & 0.829 (0.619) & 0.817 (0.636) \\
        \midrule
        GPT-4.1 (serious) & 1.101 (0.767) & 2.960 (0.648) & \textbf{3.149} (0.591) & 1.653 (0.732) & 2.082 (0.672) & 0.868 (0.645) \\
        GPT-4.1 & 2.245 (0.615) & 2.480 (0.843) & 2.594 (0.707) & 1.511 (0.623) & 1.873 (0.753) & 1.655 (0.682) \\
        Gemini 2.5 Pro & 2.294 (0.642) & 2.737 (0.710) & 2.726 (0.707) & 1.608 (0.585) & 2.077 (0.686) & 1.887 (0.740) \\
        Claude Sonnet 4 & 2.299 (0.604) & 2.313 (0.849) & 2.519 (0.780) & 1.588 (0.567) & 1.722 (0.716) & 1.456 (0.681) \\
        \bottomrule
    \end{tabular}
    \caption{Mean and standard deviation of human evaluations for each response type (image topics only)}
    \label{tab:Human_image}
\end{table}

\begin{table}[H]
    \centering
    \begin{tabular}{l|rrrrrr}
        \toprule
        & {\textbf{Novelty}} & {\textbf{Clarity}} & {\textbf{Relevance}} & {\textbf{Intelligence}} & {\textbf{Empathy}} & {\textbf{Overall Funniness}} \\
        \midrule
        High & 2.450 (0.878) & 3.820 (0.499) & 3.750 (0.655) & 1.775 (0.823) & 2.575 (0.792) & 2.615 (0.741) \\
        Mid & 2.420 (0.904) & 3.855 (0.495) & 3.745 (0.585) & 1.670 (0.833) & 2.615 (0.787) & 2.620 (0.691) \\
        Low & 2.445 (0.923) & 3.630 (0.718) & 3.625 (0.766) & 1.600 (0.891) & 2.295 (0.855) & 2.410 (0.791) \\
        Unrelated & \textbf{2.830} (0.897) & 2.705 (1.120) & 2.565 (1.105) & 1.470 (0.879) & 1.890 (1.011) & 2.315 (0.860) \\
        \midrule
        GPT-4.1 (serious) & 1.690 (1.044) & \textbf{3.995} (0.071) & 3.910 (0.416) & \textbf{1.930} (0.888) & 2.740 (0.791) & 1.850 (0.986) \\
        GPT-4.1 & 2.620 (0.734) & 3.975 (0.186) & 3.945 (0.304) & 1.820 (0.582) & 2.860 (0.730) & 2.895 (0.660) \\
        Gemini 2.5 Pro & 2.735 (0.698) & 3.985 (0.122) & 3.960 (0.262) & 1.900 (0.702) & 2.940 (0.748) & \textbf{3.010} (0.642) \\
        Claude Sonnet 4 & 2.440 (0.720) & 3.975 (0.211) & \textbf{3.970} (0.222) & 1.830 (0.658) & \textbf{2.975} (0.766) & 2.745 (0.657) \\
        \bottomrule
    \end{tabular}
    \caption{Mean and standard deviation of GPT-4.1 evaluations for each response type (text topics only)}
    \label{tab:GPT_text}
\end{table}

\begin{table}[H]
    \centering
    \begin{tabular}{l|rrrrrr}
        \toprule
        & {\textbf{Novelty}} & {\textbf{Clarity}} & {\textbf{Relevance}} & {\textbf{Intelligence}} & {\textbf{Empathy}} & {\textbf{Overall Funniness}} \\
        \midrule
        High & 2.382 (0.890) & 3.935 (0.377) & 3.688 (0.692) & 1.156 (0.835) & 2.628 (0.889) & 2.658 (0.831) \\
        Mid & 2.332 (0.943) & 3.894 (0.394) & 3.558 (0.820) & 1.161 (0.794) & 2.618 (0.896) & 2.588 (0.871) \\
        Low & 2.412 (0.985) & 3.774 (0.554) & 3.402 (0.881) & 1.151 (0.827) & 2.276 (0.858) & 2.513 (0.834) \\
        Unrelated & \textbf{2.613} (1.018) & 3.613 (0.795) & 2.683 (1.023) & 1.196 (0.845) & 2.261 (0.928) & 2.508 (0.858) \\
        \midrule
        GPT-4 (serious) & 0.698 (0.948) & \textbf{4.000} (0.000) & \textbf{3.955} (0.289) & 1.387 (0.913) & 2.759 (0.928) & 0.955 (1.079) \\
        GPT-4 & 2.427 (0.677) & \textbf{4.000} (0.000) & 3.889 (0.386) & 1.432 (0.707) & \textbf{3.050} (0.737) & \textbf{2.894} (0.598) \\
        Gemini Pro & 2.508 (0.791) & \textbf{4.000} (0.000) & 3.809 (0.563) & 1.337 (0.698) & 2.950 (0.777) & 2.884 (0.705) \\
        Claude Sonnet & 2.513 (0.822) & \textbf{4.000} (0.000) & 3.824 (0.526) & \textbf{1.548} (0.641) & 2.864 (0.743) & 2.844 (0.725) \\
        \bottomrule
    \end{tabular}
    \caption{Mean and standard deviation of GPT-4.1 evaluations for each response type (image topics only)}
    \label{tab:GPT_image}
\end{table}

\begin{table}[H]
    \centering
    \begin{tabular}{l|rrrrrr}
        \toprule
        & {\textbf{Novelty}} & {\textbf{Clarity}} & {\textbf{Relevance}} & {\textbf{Intelligence}} & {\textbf{Empathy}} & {\textbf{Overall Funniness}} \\
        \midrule
        High & 2.595 (1.076) & 3.950 (0.313) & 3.970 (0.222) & 1.180 (1.194) & 2.680 (1.206) & 3.315 (0.774) \\
        Mid & 2.605 (1.032) & 3.910 (0.378) & 3.965 (0.232) & 1.305 (1.300) & 2.750 (1.111) & 3.270 (0.707) \\
        Low & 2.690 (1.039) & 3.750 (0.728) & 3.905 (0.408) & 1.235 (1.307) & 2.380 (1.298) & 3.180 (0.819) \\
        Unrelated & \textbf{3.370} (0.858) & 2.945 (1.153) & 3.145 (1.331) & \textbf{1.605} (1.227) & 2.155 (1.404) & 3.320 (0.960) \\
        \midrule
        GPT-4.1 (serious) & 1.535 (1.276) & 3.975 (0.291) & 3.960 (0.281) & 1.085 (1.172) & 2.725 (1.244) & 1.910 (1.386) \\
        GPT-4.1 & 2.655 (0.900) & 3.965 (0.184) & 3.970 (0.264) & 0.940 (1.133) & 2.785 (1.032) & 3.215 (0.584) \\
        Gemini 2.5 Pro & 2.695 (0.840) & 3.980 (0.172) & \textbf{4.000} (0.000) & 1.090 (1.126) & \textbf{2.870} (1.029) & \textbf{3.335} (0.604) \\
        Claude Sonnet 4 & 2.385 (0.949) & \textbf{3.990} (0.100) & \textbf{4.000} (0.000) & 0.935 (1.061) & 2.860 (1.143) & 3.140 (0.642) \\
        \bottomrule
    \end{tabular}
    \caption{Mean and standard deviation of Gemini 2.5 Pro evaluations for each response type (text topics only)}
    \label{tab:Gemini_text}
\end{table}

\begin{table}[H]
    \centering
    \begin{tabular}{l|rrrrrr}
        \toprule
        & {\textbf{Novelty}} & {\textbf{Clarity}} & {\textbf{Relevance}} & {\textbf{Intelligence}} & {\textbf{Empathy}} & {\textbf{Overall Funniness}} \\
        \midrule
        High & 2.764 (0.887) & 3.930 (0.369) & 3.894 (0.572) & 1.201 (1.064) & 2.648 (1.166) & \textbf{3.412} (0.620) \\
        Mid & 2.764 (0.899) & 3.915 (0.359) & 3.910 (0.473) & 1.246 (1.139) & 2.568 (1.216) & 3.322 (0.709) \\
        Low & 2.839 (0.945) & 3.774 (0.685) & 3.839 (0.581) & 1.392 (1.158) & 2.246 (1.170) & 3.296 (0.665) \\
        Unrelated & \textbf{3.116} (0.712) & 3.588 (0.911) & 3.256 (1.141) & \textbf{1.588} (1.146) & 2.216 (1.314) & 3.261 (0.906) \\
        \midrule
        GPT-4.1 (serious) & 0.548 (1.018) & 3.985 (0.213) & 3.950 (0.386) & 0.472 (0.834) & 1.709 (1.469) & 0.754 (1.212) \\
        GPT-4.1 & 2.472 (0.834) & 3.990 (0.142) & 3.935 (0.334) & 0.975 (1.017) & 2.799 (1.092) & 3.191 (0.598) \\
        Gemini 2.5 Pro & 2.508 (0.898) & \textbf{3.995} (0.071) & \textbf{3.975} (0.186) & 0.844 (1.050) & \textbf{2.889} (1.105) & 3.357 (0.576) \\
        Claude Sonnet 4 & 2.533 (0.809) & 3.970 (0.244) & 3.894 (0.465) & 1.161 (1.061) & 2.693 (1.074) & 3.171 (0.690) \\
        \bottomrule
    \end{tabular}
    \caption{Mean and standard deviation of Gemini 2.5 Pro evaluations for each response type (image topic only)}
    \label{tab:Gemini_image}
\end{table}

\begin{table}[H]
    \centering
    \begin{tabular}{l|rrrrrr}
        \toprule
        & {\textbf{Novelty}} & {\textbf{Clarity}} & {\textbf{Relevance}} & {\textbf{Intelligence}} & {\textbf{Empathy}} & {\textbf{Overall Funniness}} \\
        \midrule
        High & 2.915 (0.671) & 3.820 (0.591) & 3.840 (0.580) & 2.060 (0.706) & 2.920 (0.629) & 2.945 (0.595) \\
        Mid & 2.815 (0.777) & 3.740 (0.752) & 3.745 (0.743) & \textbf{2.095} (0.806) & 2.810 (0.739) & 2.820 (0.714) \\
        Low & 2.725 (0.891) & 3.450 (1.021) & 3.505 (1.017) & 1.820 (0.837) & 2.620 (0.871) & 2.610 (0.884) \\
        Unrelated & 2.715 (1.140) & 2.165 (1.428) & 2.055 (1.528) & 1.610 (1.016) & 1.740 (1.257) & 1.990 (1.182) \\
        \midrule
        GPT-4.1 (serious) & 1.745 (1.165) & 3.960 (0.314) & 3.800 (0.702) & 1.675 (0.844) & 2.695 (0.797) & 1.825 (1.123) \\
        GPT-4.1 & 2.860 (0.642) & 3.960 (0.281) & 3.940 (0.356) & 1.890 (0.538) & 3.005 (0.465) & 2.925 (0.530) \\
        Gemini 2.5 Pro & \textbf{2.935} (0.541) & 3.980 (0.223) & \textbf{3.990} (0.141) & 2.050 (0.565) & \textbf{3.095} (0.487) & \textbf{3.040} (0.447) \\
        Claude Sonnet 4 & 2.755 (0.705) & \textbf{3.995} (0.071) & 3.985 (0.158) & 1.970 (0.584) & 3.030 (0.539) & 2.940 (0.498) \\
        \bottomrule
    \end{tabular}
    \caption{Mean and standard deviation of Claude Sonnet 4 evaluations for each response type (text topics only)}
    \label{tab:Claude_text}
\end{table}

\begin{table}[H]
    \centering
    \begin{tabular}{l|rrrrrr}
        \toprule
        & {\textbf{Novelty}} & {\textbf{Clarity}} & {\textbf{Relevance}} & {\textbf{Intelligence}} & {\textbf{Empathy}} & {\textbf{Overall Funniness}} \\
        \midrule
        High & 2.578 (0.812) & 3.809 (0.662) & 3.658 (0.794) & 1.678 (0.730) & 2.759 (0.760) & 2.608 (0.723) \\
        Mid & 2.462 (0.903) & 3.558 (0.962) & 3.472 (1.063) & 1.548 (0.795) & 2.593 (0.910) & 2.407 (0.859) \\
        Low & 2.538 (0.851) & 3.513 (1.014) & 3.508 (1.044) & 1.603 (0.790) & 2.523 (0.881) & 2.442 (0.807) \\
        Unrelated & \textbf{2.683} (0.850) & 3.176 (1.253) & 2.874 (1.477) & 1.704 (0.750) & 2.312 (1.007) & 2.377 (0.923) \\
        \midrule
        GPT-4.1 (serious) & 0.779 (1.055) & 3.940 (0.434) & 2.955 (1.405) & 0.754 (0.844) & 1.804 (1.162) & 0.834 (1.077) \\
        GPT-4.1 & 2.658 (0.623) & \textbf{4.000} (0.000) & 3.905 (0.397) & \textbf{1.794} (0.525) & \textbf{3.025} (0.497) & \textbf{2.764} (0.460) \\
        Gemini 2.5 Pro & 2.643 (0.717) & 3.985 (0.213) & 3.819 (0.617) & 1.618 (0.607) & 2.930 (0.564) & 2.744 (0.541) \\
        Claude Sonnet 4 & \textbf{2.683} (0.640) & 3.980 (0.200) & \textbf{3.945} (0.365) & 1.754 (0.546) & 2.879 (0.537) & 2.719 (0.561) \\
        \bottomrule
    \end{tabular}
    \caption{Mean and standard deviation of Claude Sonnet 4 evaluations for each response type (image topics only)}
    \label{tab:Claude_image}
\end{table}

\end{document}